\documentclass{elsarticle}
\usepackage{amsthm}
\usepackage{amsmath,amssymb}
\newtheorem{definition}{Definition}[section] 
\usepackage{lineno,hyperref}
\usepackage{graphicx}%
\usepackage{ulem}
\usepackage{epsfig}
\usepackage{amsfonts}
\newtheorem{theorem}{Theorem}
\usepackage{xcolor}
\newcommand{\eps}{\varepsilon} %GH's
\usepackage{makeidx}         % allows index generation
\usepackage{graphicx}   
\usepackage{subcaption}

\newcommand{\real}{\mathbb{R}}

\newcommand{\realpx}{\real^{\rm{p+1}}}

\newcommand{\x}{\mathbb{X}}
\newcommand{\y}{\mathbb{Y}}
\newcommand{\nn}{\mathbb{N}}
\newcommand{\K}{\mathbb{K}}
 \usepackage{cleveref}
\usepackage{float}

\newcommand{\vc}[1]{\mathbf{#1}}
\newcommand{\vtc}[1]{\mathbf{\tilde{#1}}}

\newcommand{\mb}[1]{{\boldsymbol #1}}
\newcommand{\al}{\alpha}
\newcommand{\mba}{\mb{\al}}
\newcommand{\mbu}{\mb{\mu}}
\newcommand{\Beta}{\mathrm{B}}

\makeatletter
\def\ps@pprintTitle{%
	\let\@oddhead\@empty
	\let\@evenhead\@empty
	\let\@oddfoot\@empty
	\let\@evenfoot\@oddfoot
}
\makeatother
\begin{document}

\begin{frontmatter}
\title{Phenotyping OSA: a time series analysis using fuzzy clustering and persistent homology}
%\title{Clustering Time Series}
%\tnotetext[mytitlenote]{Fully documented templates are available in the elsarticle package on \href{http://www.ctan.org/tex-archive/macros/latex/contrib/elsarticle}{CTAN}.
%}

%\author{Authors}
\maketitle

\author[mymainaddress]{Prachi Loliencar}
\ead{lolienca@ualberta.ca}

\author[mymainaddress]{Giseon Heo\corref{mycorrespondingauthor}}
\cortext[mycorrespondingauthor]{Corresponding author}
\ead{gheo@ualberta.ca}

\address[mymainaddress]{School of Dentistry; Department of Mathematical and Statistical Sciences, University of Alberta, Edmonton Alberta, T6G 1C9, Canada}
%\fi
\begin{abstract}
Sleep apnea is a disorder that has serious consequences for the pediatric population. There has been recent concern that traditional diagnosis of the disorder using the apnea-hypopnea index may be ineffective in capturing its multi-faceted outcomes. In this work, we take a first step in addressing this issue by phenotyping patients using a clustering analysis of airflow time series. This is approached in three ways: using feature-based fuzzy clustering in the time and frequency domains, and using persistent homology to study the signal from a topological perspective. The fuzzy clusters are analyzed in a novel manner using a Dirichlet regression analysis, while the topological approach leverages Takens embedding theorem to study the periodicity properties of the signals.
\end{abstract}


\end{frontmatter}

%\linenumbers

\section{Introduction} 

Obstructive sleep apnea (OSA) is a sleep disorder involving partial or complete airway collapse or obstruction, causing breathing issues and disruptions in normal sleep patterns. It has serious consequences for the pediatric population, affecting not only their health but also their neurodevelopment. The gold standard for diagnosis is overnight polysomnography~(PSG), a procedure that is expensive, and hard to access in many countries. The severity of OSA is diagnosed using the the apnea-hypopnea index ($ahi$)- the average number of apnea (little to no airflow) and hypopnea (partial airflow) events the patient underwent per hour of sleep.  To identify these events, sleep technicians primarily use airflow signals, which are recorded as temperature signals during the PSG. 
%OSA severity in children and adolescents is then categorized as follows: {low risk} ($\text{\emph{ahi}}<2$), {mild} ($ 2 \leq\text{\emph{ahi}}\leq 5$), {moderate} ( $5 < \text{\emph{ahi}}\leq 10$), and {severe} ($\text{\emph{ahi}}>10$).  
Recently there have been concerns in the medical community regarding the simplistic nature of using a single metric such as $ahi$, in diagnosing complex, multi-faceted disorders like OSA \cite{ditchAHI2020}. Additionally, the $ahi$ has not proven to be effective in helping to choose optimal treatments for patients \cite{Rosen2015}. In this paper, we hope to take a starting step in using machine learning to understand OSA without relying on the $ahi$.

%In the pediatric population especially, this can not only have health consequences but also affect the child's physical and neurological development. 

There have been many studies on predicting OSA and its outcomes using clinical covariates and time series data \cite{Mostafa2019}. However, these rely on patient labels identified using $ahi$, and therefore provide no additional information on the disorder. Researchers have recently taken up using unsupervised methods, particularly clustering, to identify subtypes or phenotypes of OSA, with the idea that studying the patient data directly may reveal interesting patterns that are not captured by the $ahi$ \cite{Zinchuk2020}. As far as we know, none of these have used features directly obtained from the timeseries recorded during PSG. As such, our goal in this paper is to conduct a clustering and exploratory analysis of OSA using airflow signals, with the objective of identifying phenotypes.

Seventy four pediatric patients with potential OSA were recruited in a clinical study (Pro00057638, approved by University of Alberta). These patients underwent an overnight PSG and filled out several survey questionnaires providing the data that forms the basis of this study.  In \cite{loliencarPSG2021}, we conducted observation-based clustering on the airflow signals and sleep stages of these patients using the dynamic time warping distance. In this work, we focus instead on feature-based clustering of airflow signals using three different perspectives: in the time domain using the partial autocorrelation function, in the frequency domain using spectral density estimates, and topologically using persistent homology. For the first two cases, we use fuzzy clustering using medoids, then follow this up with a Dirichlet regression analysis, while the latter is followed up with LDA.

Fuzzy clustering outputs probability vectors for each data point, indicating the degrees of membership of that point to each cluster. Most classification algorithms are unable to handle such inputs for the dependent variable and require hard classes. Dirichlet regression \cite{hijazi1, campbell1, maier} on the other hand, allows the parameters of the Dirichlet distribution to be modelled on covariates, and is thereby designed to handle compositional data. We therefore use this technique in a novel manner to find the covariates that are most significant in modelling our fuzzy clusters. 
 
Topological data analysis (TDA) is an emerging research area in data science that has been used in diverse applications \cite{zhuPH-AI, biologyCite, signalCite2,GIDEA2018820}. Its foundation lies in a topological method called \emph{persistent homology} which quantifies the topological features of shapes and functions. In particular, the method involves constructing filtered simplicial complexes on the dataset and measuring the evolution of topological features over changes in the filtration parameter \cite{comp-top,ZomorodianCPH} for a given dimension.  To apply TDA to airflow signals, we first convert each time series into a point cloud using a time-delay embedding, an example of which is Takens' delay embedding  \cite{Takens1981}.  The embedding picks up repeating patterns in the time series which is translated into periodicity in the generated point cloud \cite{perea15A, perea15B}. This periodicity is used to obtain a Morse function (see \cref{sec:ph}) for each airflow signal, and we use 0-dimensional persistence homology to obtain a complete-linkage hierarchical clustering of the data. We follow this up with identifying phenotypes using a linear discriminant analysis. For general references on TDA and its applications to timeseries, we refer the reader to \cite{kim2019time,ravishanker2019topological,gholizadeh2018short}.

Our article is organised as follows: we introduce fuzzy clustering and Dirichlet regression in \cref{sec:fuzzy} and \cref{sec:dirich} respectively, followed by introduction to persistent homology in \cref{sec:ph}; we introduce our data and covariates in \cref{sec:data}; we then present the results of phenotyping using fuzzy clustering in \cref{sec:phenotyping}, and the results of a TDA analysis in  \cref{sec:tda_analysis}. Finally, we conclude with \cref{sec:conclusion} with a summary of the paper and directions for future work.

\section{Fuzzy Clustering}
\label{sec:fuzzy}
Fuzzy clustering refers to a class of unsupervised learning methods generalizing $K$-means and $K$-medoids clustering, by allowing a data point to be a member to multiple clusters. In particular, each data point is assigned a $K$-dimensional probability vector whose $i$-th component gives the probability of the point belonging to the $i$-th cluster. This allows a lot of flexibility compared to hard clustering methods. For example, in clinical applications, as in our case, a patient may exhibit multiple phenotypes (clusters) with similar probabilities. Instead of breaking ties randomly, or discarding the second most likely phenotype by a small margin, fuzzy clustering can allow the clinician to assess the different possibilities. This includes risk monitoring if the patient has a reasonable probability of exhibiting more severe symptoms.

For our application, we will use $K$ medoids - representatives of clusters picked from the original dataset. This ensures that we do not use cluster representatives (such as centroids) that may not exist in the real world, or that do not make sense from a clinical perspective. Given a set of $n$ data points, $\vc{X}=\{\vc{x}_1,\vc{x}_2,...,\vc{x}_n\}$, in $\mathbb{R}^p$, fuzzy clustering with medoids optimizes the following objective: 
\begin{equation}
\label{eqn:fuzzyobj}
\begin{aligned}
& \underset{\{\vtc{x}_c\}_{c=1}^K, \{u_{ic}\}_{c=1}^K, i\in [n]}{\min}
& & \sum_{i=1}^n\sum_{c=1}^K u_{ic}^m \ d(\vc{x}_i,\vtc{x}_c)   \\
& \text{subject to}	& & \sum_{c=1}^K u_{ic}=1, \ u_{ic}\geq 0,\\ & \  & & \text{ for } c\in [K], i\in[n] \\
%	& \text{ }
%	& & \xi_i\geq 0, \;    \text{ for } i = 1, \ldots, m.
\end{aligned}
\end{equation}

\noindent where $\{\vtc{x}_1,\vtc{x}_2,...,\vtc{x}_K\}\subset \vc{X}$ denote the representative $K$-medoids, the  $u_{ic}$ terms indicate the degree of membership of the $i$th data point to the $c$th cluster, and  $m\in\mathbb{N}$ is a hyperparameter that controls the degree of fuzziness ($m=1$ gives hard clusters while all clusters become equally probable as $m\to \infty$).  We use $m=1.5$ as our fuzziness parameter as recommended in section 4.4 of \cite{Maharaj2019}. The objective \ref{eqn:fuzzyobj} can then be optimized using an iterative algorithm called the Fuzzy C-medoids algorithm proposed by Krishnapuram in \cite{krishnapuram}. For our implementation, we use the \texttt{dtwclust} package in \texttt{R} \cite{sarda}. The interested reader can learn more about fuzzy clustering in \cite{Maharaj2019}.

\section{Dirichlet Regression}
\label{sec:dirich}
Most supervised learning methods for classification, such as logistic regression, are only equipped to handle absolute categorical inputs for the dependent variable, i.e. the class probabilities for each point must be 0 or 1. These methods are unable to deal with compositional data or with instances where the data point belongs to multiple classes to varying degrees, an example being the output of the fuzzy clustering class of algorithms. To accommodate this, we use Dirichlet Regression using the \texttt{DirichletReg} package \cite{dirichpack}.

The Dirichlet distribution, best known as the conjugate prior of the multinomial distribution, models the probability mass function over a $(K-1)$-simplex, i.e. over possible probability vectors of $K$ events, parametrized by a vector of positive reals. This is a generalization of the Beta distribution to more than two components. In particular, given $K\geq 2$ categories and $\mb{\al}=(\al_1,\al_2,...,\al_K)$ with $\al_i>0$ for all $1\leq i\leq K$, the probability density function of the Dirichlet distribution $\text{Dir}(\al)$ over the $K-1$ simplex is given by

\begin{equation}
f(\vc{y}; \mb{\al}) = \frac{1}{\Beta(\mba)} \prod_{c=1}^K y_c^{\al_c-1} 
\label{dirich}
\end{equation}
where $\vc{y}=(y_1,...,y_{K-1})$ is in the $(K-1)$-simplex i.e. is a vector of positive reals satisfying $\sum_{i=1}^K y_i=1$. Here $\Beta(\cdot)$ refers to the multivariate Beta function, given by $\Beta(\mba)= {\prod_{c=1}^K\Gamma(\al_c)}/{\Gamma\left(\sum_{c=1}^K \al_c\right)}$ with respect to the Gamma function. This term serves as a normalization constant in \cref{dirich}. Setting $\al_0=\sum_{c=1}^K \al_c$, the marginal expected values and corresponding variances of the variables in the distribution are given by $\mathrm{E}[y_c|\mba]= \frac{\al_c}{\al_0}$, $\mathrm{Var}[y_c|\mba]= \frac{\al_c(\al_0-\al_c)}{\al_0^2(\al_0+1)}$. Additionally, covariance between two variables is given by $\mathrm{Cov}[y_i,y_j|\mba]=-\frac{\al_i\al_j}{\al_0^2(\al_0+1)}$.

In this work we use an alternate parametrization introduced by Ferrari and Cribari-Neto for the beta distribution in \cite{betareg1,betareg2} and extended to the
Dirichlet distribution by Maier in \cite{maier}. This parametrization uses the variables $\mu_c=E[y_c|\mba]$ and $\phi=\al_0=\sum_{c=1}^K \al_c$, so that the probability density function can be rewritten as 
\begin{equation*}
f(\vc{y}; \mbu, \phi ) = \frac{1}{\Beta(\phi\mbu)} \prod_{c=1}^K y_c^{\phi\mu_c-1} 
\end{equation*}
where $\phi>0$ and $\mbu=(\mu_1,...,\mu_K)$ satisfies $\mu_c>0$ and $\sum_{c=1}^K \mu_c=1$. The interpretation of $\mbu$ is clear as being the vector of means corresponding to $\vc{y}$. The parameter $\phi$ on the other hand, called the precision parameter, regulates how the density is ``spread out". Low precision values correspond to peaks in the density at the corners of the simplex, i.e. extreme values of 0 or 1 for one or more components of $\vc{y}$, while high values cause the distribution to be centered (unimodal) at $\mbu$.  In terms of $\mba$, we note that $\mba=\phi\mbu$ with values below 1 correspond to the first case, while values greater than 1 correspond to the second case. Lastly, $\mba=(1,1..,1)$ corresponds to a marginal uniform distribution. Visualizations of these can be found in figure 2 of \cite{maier}.

In order to model compositional data, Campbell and Mosimann introduced Dirichlet regression, allowing the parameters of the Dirichlet distribution to be dependent on a set of covariates \cite{campbell1,campbell2}. An application of this model to psychiatric data can be found in \cite{pyschdirich}.
 The residual properties and modelling of compostional data of these models were further explored by Hijazi et al in \cite{hijazi1,hijazi2}. 
 We use the version of the model presented by Maier in \cite{maier}, modelling $\mbu$ using
\begin{align*}
g_{\mu}(\mu_c)=\vc{X}{\mb{\beta}_c} \text{ for all } c\in\{1,...,K\}
\end{align*}
 where $g_\mu$ is the logit link function, $\vc{X}$ is the vector of covariates concatenated with 1 for the bias term and $\mb{\beta}_c$ are the corresponding linear coefficients shared by all $c\in \{1,...,K\}$. The parameter $\phi$ may also chosen to be dependent on covariates, however, for simplicity, we choose it to be modelled by a constant to be determined by the data. The regression is fitted using a maximum log likelihood cost function and several optimization algorithms, details of which and more can be found in \cite{maier}. 

In our study, we use Dirichlet regression to model the outputs of our Fuzzy clustering algorithm, using covariates described in \cref{sec:data}. We covariates with significant $p$-values, in order to understand the underlying phenotypes detected by the clustering.

\section{Topological data analysis for point cloud and Morse function}\label{sec:ph}

We begin by summarizing persistent homology on a point cloud, and then do the same for a Morse function (a smooth function on the manifold with non-degenerate critical points).

Consider a finite set of data points $X=\{X_1, X_2, \ldots, X_n\}$ embedded in a $p$-dimensional metric space $\y$, with metric $d: \y \times \y \rightarrow \real$. We assume that this data is sampled from an unknown lower dimensional subspace $\x$ embedded topologically in $\y$.  Due to sampling, $X$ does not capture the the geometry and topology of $\x$, and we rely on building simplicial complexes on $X$ to recover the topological properties of the underlying space $\x$. Here, given a set $S$, an (abstract) simplicial complex is defined to be a set $K$, of finite subsets of $S$ such that $\alpha \in K$ and $\alpha' \subset \alpha$ implies $\alpha'\in K$. Two common choices of simplicial complexes are the {\v Cech} complex and the Vietoris-Rips complex, defined as follows:
\begin{definition}%{\bf {\v Cech} complex}\;}
Given $\eps>0$, the \emph{\v Cech} complex, $C_X(\eps)$, is the simplicial complex given by
$$C_X(\eps) := \{ \sigma\subset X: \bigcap_{x\in \sigma} B_{\y}(x, \eps) \neq \emptyset \}$$
where $B_{\y}(x,\eps)= \{ y \in \y \;|\; d(x,y) \leq \eps\}$ denotes the closed ball of radius $\eps >0$ centered at $x$.
\end{definition}
In other words, a collection of points $\sigma=(x_0, \ldots, x_p)$ forms a $p$-simplex if the set of balls of radius $\eps$ centered at these points has non-empty intersection.
\begin{definition}
Given $\eps>0$, the Vietoris-Rips complex, $V_X(\eps)$ is defined by
$$V_X(\eps):=\{\sigma \subset X:\, d(x_i, x_j) \leq 2\eps, \forall x_i, x_j \in \sigma\}$$
\end{definition}
The  {\v Cech} complex and the Vietoris-complex have a chain of inclusion, $V_X(\eps) \subset C_X(2\eps) \subset V_X(2\eps)$, \cite{deSilvaConvg}. The following Nerve Theorem (\cite{rotman} and \cite{mccord})  connects the \emph{\v Cech} complex and underlying continuous space.
\begin{theorem}
Given $\{x_1, ..., x_k\} \subset X$, and $\eps >0$, if $\bigcap^k_{i=i} B_{\y} (x_i, \eps)$ is either empty or contractible, then the {\v Cech complex}, $C_X(\eps)$  is homotopy equivalent to the unions of balls $\bigcup _{x\in X}B_{\y}(x, \eps).$
\end{theorem}
The {\v Cech} complex is massive at large scales and computationally expensive,  so in practice, we use other complexes. The Vietoris-Rips complex is most popular due to its simple construction, even for higher dimensions, and will be used in this paper.

One can find $\eps>0$ such that the complex of set of sample points, $X$ is homotopy equivalent to the union of the balls, $\bigcup _{x\in X}B_{\y}(x, \eps),$ and thus they have the same homology. Instead of finding the one value of $\eps>0$ that captures the topology of $X$, computational topologists developed \emph{persistent homology} to reveal persistent topological features as $\eps$ is varied. In other words, persistent homology studies the way topological features evolve over different resolutions. For a comprehensive survey of persistent homology, refer to  \cite{comp-top, ZomorodianCPH}. 

We now briefly summarize the mathematical definition of persistent homology. 
\begin{definition}
A filtration $\K=\{K_a\}_{a\in \real}$ is s collection of simplicial complexes $K_a$ of $X$ for which $a \leq b$ implies $K_a \subset K_b$ (see \cref{fig:ripsfilt} for a pictorial representation of the Vietoris-Rips filtration).

\end{definition}
\begin{definition}
For each $a\in \mathbb{R}$, let $H_p\K_a$ be the $p$-dimensional homology of $\K_a,$ where $p \in \nn$. The associated $p$-dimensional  persistent homology $\mathcal{P}H_p$ is a collection of vector spaces $\{H_p K_a\}_{a \in \real}$ equipped with homomorphisms 
$\{\iota^{a,b}_p\},$ where
$\iota_p^{a,b}: H_p(K_a) \rightarrow H_p(K_b)$ is the homomorphism induced by the inclusion $K_a \subset K_b.$
\end{definition}
Corresponding to $p$-dimensional persistent homology, a \emph{barcode} \cite{barcodeZomorodian} is a finite multiset consisting of intervals $[b, d)$ of filtration values, where for a given  homology, $b$ and $d$ indicates its appearance and disappearance in $\mathcal{P}H_p$, and are known as the homology's time of birth and death respectively (see \cref{fig:ripsfilt} in the Appendix). 
These birth and death times can also be plotted in 2-D as $(x,y)$ pairs and the representation of the persistent homology in this form is called a \emph{persistence diagram} \cite{edels2002} (examples of persistence diagrams are provided in \cref{fig:exTypical} in the appendix). 
One can then measure dissimilarity between persistence diagrams using bottleneck distance.
\begin{definition} 
Let $\rm{dgm}_p(K)$ and $\rm{dgm}_p(K'),$ $p$-dimensional persistence diagrams (including points on the diagonal) corresponding to homologies $\mathcal{P}H_p(K)$ and $\mathcal{P}H_p(K')$. Let $\Gamma$ denote the collection of all bijections $ \gamma_p:\rm{dgm}_p(K) \rightarrow \rm{dgm}_p(K')$. Then, The bottleneck distance between two persistence diagrams is defined as 
\begin{equation}\label{eqn:bottleneck}
d_B(\rm{dgm}_p(K), \rm{dgm}_p(K'))= \inf_{\gamma_p \in \Gamma} \sup_{x\in \rm{dgm}_p(K)} ||x-\gamma_p(x)||_{\infty}.
\end{equation}
\end{definition}

In computational topology, the modality of a Morse function is observed by looking at the connectivity of its support as the height of the function decreases. More precisely, consider the super-level sets 
$ \real_h=\{x\in \real\mid f(x) \geq h\},$ for each $h\in \real.$
As we decrease $h$, the connectivity of $\real_h$ remains the same except when we pass a critical value.
 At a local maximum, the super-level set adds a new component, while at a local minimum, two components merge into one.
We pair a local minimum with the lower of the two local maxima that represent the two components.
 The other maximum is the representative of the component resulting from the merger.
 When $x_1$ and $x_2$ are paired in this manner, we define the persistence of the pair to be $f(x_2)-f(x_1)$.
Persistence is visualized in the persistence diagram by mapping each pair to the point $(f(x_1), f(x_2))$. We present an illustration of super-level set filtration in Figure \ref{fig:morseF} in the Appendix.

TDA is effective in detecting useful topological features in point cloud data \cite{carlsson_2014}. Our goal is to convert a time series to a point cloud and analyse it by persistent homology. Motivated by work of \cite{perea15A, perea15B, kim2019time}, we apply sliding window embedding in our analysis.
\begin{definition}
Given a time series $X(t), t \in [0, T], 0 < T < \infty,$
the sliding window embedding is a map  $\phi: \real \rightarrow \realpx,$
$\phi X(t)=[X(t), X(t+\tau), \ldots, X(t+p\tau)],\,$ where $\, \tau\in \real$ is the delay parameter and  $p\in \nn$ is embedding dimension.
\end{definition}
 The collection of values at different $t\in [0,T]$ produces a point cloud. Denote a sequence of samples $\{x(t_0), x(t_1), \ldots, x(t_n)\}$ taken at each time point, $0=t_0, t_1, \ldots, t_{n-1}, t_n=T.$  We represent a point cloud as a trajectory matrix $X\in \real^{(n-p\tau)\times (p+1)},$ as follows
\begin{equation}\label{eqn:trajec}
X=
\begin{bmatrix}
\phi X(t_0) \\
\phi X(t_1)\\
\vdots\\
\phi X(t_n)
\end{bmatrix}
=
\begin{bmatrix}
x(t_0) & x(t_{\tau}) & \cdots & x(x_{p\tau})\\
x(t_1) & x(t_{1+\tau}) & \cdots & x(t_{1+p\tau}) \\
\vdots & \vdots  &  & \vdots \\
x(t_{n+p\tau}) & x(t_{n+(p+1)\tau})& \cdots & x(t_n)
\end{bmatrix}
\end{equation}

The sequence $\phi X(t)$ traces a closed curve in $\real^{p+1}$ which is 1-dimensional hole~(loop), thus its persistence measures the degree of circularity of the point cloud. The maximum 1D persistence occurs when the \emph{window-size}, $p\times \tau$, corresponds to the frequency of the signal $X(t).$   As a measure of periodicity, the authors \cite{perea15B}, recommended $(d^n-b^m)/3^{n/2},$ where $b-d$ is the largest persistence in one dimensional homological feature, and called it \emph{periodicity score}. In our application, we choose $n=m=1.$ 

Takens' theorem proves that there exists a pair of parameters $(\tau, p)$ such that the embedding preserves the underlying structure of the topological space. The embedding is sensitive to the choices of the parameters, $p$ and $\tau$. 
There is no generic optimal method for choices of these two parameters.  In general, it is recommended for data scientists to choose large $p$, and  $\tau$ based on autocorrelation such that the lag at its autocorrelation is $<2/\sqrt{T}.$  Another challenge is in choosing $n,$  circular shapes may have no chance to be formed due to not enough number of points, while too many points likely causes high computational time.

Since the embedding dimension $p$ is set to be large, we may run into the `curse of dimensionality', and real world data may contain outliers and noise. Kim \emph{et al} \cite{kim2019time} suggested applying principal component analysis~(PCA) to address these issues. The authors show that applying PCA reduces high dimensional topological noise, at the same time, preserving  important topological features. For examples of sliding window embedding with synthetic time series, see Figure \ref{fig:embedding} in the Appendix.

\section{Data and covariates}\label{sec:data}
 \Cref{tab:class-psg} summarizes the distribution of OSA severities in our 74 pediatric patients, based on the $ahi$ metric.

\begin{table}[h]
	\small
	\centering
	\renewcommand{\arraystretch}{1.2}\resizebox{\textwidth}{!}{
	\begin{tabular}{|c|c|c|c|c|}
		\hline
		{\bf Class}& Low risk &  Mild OSA & Moderate OSA & Severe OSA\\ \hline     
		{\bf \emph{ahi} score} & ${\emph{ahi}} <2$ & $ 2 \leq {\emph{ahi}} \leq 5$ & $ 5 < {\emph{ahi}} \leq 10$  &${\emph{ahi}} >10$  \\ \hline
		{\bf Number of subjects} & 20~(27\%) & 26~(35\%) & 14~(19\%) &14~(19\%) \\ \hline
		{\bf Avg. Period score} & 0.63 $\pm$ 0.054 & 0.63 $\pm$ 0.09 & 0.57 $\pm$ 0.12 & 0.50$\pm$ 0.11 \\ \hline
	\end{tabular}}
	\caption{Distribution of OSA patients based on an \emph{ahi}-based classification. The second row indicates the cut off value of \emph{ahi} scores for each group. The last row presents the average of periodicity scores and standard deviation (see \cref{sec:tda_analysis}).}
	\label{tab:class-psg}
\end{table}

These patients underwent overnight PSG during which airflow and various brain signals such as EEG, ECG and EMG were recorded. A sleep technician conducted sleep staging for each patient using the brain signals, labelling each 30s epoch into one of the five states: wake, rapid eye movement (REM), and three non-rapid eye movements (NREM1, NREM2, NREM3). We computed the proportion of sleep states corresponding to each of these labels and used these as covariates (in our analysis we omitted REM as the variables are correlated with sum 1). The airflow signal was sampled at 32Hz, and we used a fourth-order low-pass forward-backward Butterworth filter with a cutoff of 1.2Hz to preprocess the signal.

In addition to the PSG, the participating patients filled out the following questionnaires (possibly with the help of a guardian):  Child Sleep Habits Questionnaire~(CSHQ), OSA Quality of Life~(OSA-18), Pediatric Quality of Life Inventory~(PedsQL), and Pediatric sleep questionnaire~(PSQ). Each questionnaire consists of a set of questions regarding frequency or severity of symptoms and requires Likert-style responses. Several of these are divided into categories covering symptom types, such as: daytime sleepiness issues, nighttime sleep disturbances, physical suffering, emotional problems and performance in school (school functioning). The number of questions (response types) were 45 in CSHQ (Likert scale 1-3), 18 in OSA-18 (Likert scale 1-7), 18 in PedsQL (Likert scale 0-4), and 22 in PSQ (yes or no). Higher levels on the Likert scale indicate worse symptoms with the exception of the quality of life rating. To obtain variables, we sum up the Likert responses in each symptom category, for each questionnaire, summarizing all the questions by 30 covariates. The exception to this included a few questions requiring numerical responses, inquiring about how long the child's nighttime waking lasts, or the number of hours the child usually sleeps in a day. We also included 2 additional survey responses to frequency of exercise and physical activity.
For details on the categories and types of questions, we refer the reader to \cite{cshq, osa18, pedsql, psq}. 
Lastly, the patients had their craniofacial index (CFI) evaluated by a dentist. The CFI is a sum of eight facial and dental characteristics, each of which are scored between 0,1 or 2 with higher values indicating a larger degree of abnormality.

We summarize the variables that were the most significant in our analysis in \cref{tab:variables}. The first column presents the survey name, and the second gives the variable name we assigned to the responses in a given category. The third column indicates the possible min and max of the sum of Likert responses - for example, osa-daytime consisted of 3 questions with responses between 1 and 7 based on frequency of symptoms of daytime sleepiness, so that the sum of responses fall between 3 and 21. The last column gives a brief description of the types of questions whose responses comprise the variable.

\begin{table}
\label{tab:data}
	\small
	\renewcommand{\arraystretch}{1.2}
	\begin{tabular}{|p{1cm}|p{2.5cm}|p{1.3cm}|p{6cm}|}
		\hline
		{\bf Source}&	{\bf Variable Name} & {\bf min/max} &  {\bf Interpretation} \\ \hline
		&cshq-day-sleep & (15, 45) & {sum of responses on daytime sleepiness}\\
		\cline{2-4}
		CSHQ	&cshq-night-sleep & (35, 105) & {sum of responses on nighttime sleep issues}\\  
		\cline{2-4}
		&cshq-night-waking &  & {number of minutes child's night waking usually lasts}\\ 
		\cline{2-4}
		&cshq-sleep-hrs&   & {number of hours child usually sleeps in a day (including naps)}\\ 
		\cline{2-4} 
		\hline
		&osa-daytime& (3, 21)  & {sum of responses on symptoms of daytime sleepiness}\\ 
		\cline{2-4} 
		&osa-sleep-dist& (4, 28) & {sum of responses on nighttime sleep disturbances}\\ 
		\cline{2-4} 
		&osa-phys-suff&  (4, 28) & {sum of responses on symptoms causing physical suffering}\\ 
		\cline{2-4} 
		OSA-18	&osa-emotion-stress & (3, 21)& {sum of responses on emotional stress/issues}\\
		\cline{2-4}
		&osa-caregiver& (4, 28) &{sum of responses on caregiver's concerns for child} \\
		\cline{2-4}
		&osa-overall& (0, 10) & {score of overall quality of life given by guardian} \\
		\hline
		&pedsql-emotional-func & (0, 20) & {sum of responses on emotional functioning}\\
		\cline{2-4}
		&pedsql-phys-func& (0, 32) & {sum of responses on physical functioning}\\
		\cline{2-4}
		PedsQL	&pedsql-school-func & (0, 20)& {sum of responses on functioning at school}\\
		\cline{2-4}
		&pedsql-social-func& (0, 20) & {sum of responses  on social functioning}\\
		\hline
		PSQ 	&psq &  (0, 22) &{sum of ``yes'' responses to the PSQ questionnaire}\\
		\hline
		General \hspace{1cm} survey  &physical-activity & & {number days in the past week child has participated in some form of physical activity}\\ 
		%			\cline{2-3}	
		%	General	&cfi & {craniofacial index (sum of 8 craniofacial variables)} \\
		\cline{2-4}
		& exercise & &{number of hours per week the child usually exercises}\\ %\hline
		
		%		Dental	&cfi & {craniofacial index (sum of 8 craniofacial variables)} \\
		\hline
		%		&NREM1(\%) & {percentage of sleep stages that are NREM1}\\ 
		%		\cline{2-3}
		&NREM2(\%) & & {percentage of sleep stages that are NREM2}\\
		\cline{2-4}
		%		&NREM3 (\%) & {percentage of sleep stages that are NREM3}\\
		\cline{2-4}
		PSG			&Wake (\%) & & {percentage of sleep stages that are Wake}\\
		\cline{2-4}
		%		&total-sleep & {total amount of sleep during PSG in hours }\\
		%		\cline{2-3}
		%PSG		&total-arousal-index  &  {total number of arousals* per hour during PSG  } \\
		\cline{2-4}
		features	&total-sleep-awakenings & & {total number of awakenings during PSG } \\
		\cline{2-4}
		%	&total-sleep-stage-changes &   {number of transitions from one sleep state to another}\\
		\cline{2-4}
		&sleep-efficiency(\%) & & {ratio of total sleep time and total PSG recording time} \\
		%	\cline{2-3}
		%	&waso & {wake after sleep onset in minutes; this is the total PSG recording time minus sleep latency and minus time spent asleep}\\
		\hline 
	\end{tabular}
	\caption{Description of the most significant variables in our study (see \cref{sec:data} for details). The (min, max) column shows the possible min and max of each variable (not the actual min/max in our data).}
	\label{tab:variables}
\end{table}

\section{Phenotyping using fuzzy clustering and Dirichlet regression}
\label{sec:phenotyping}

Since airflow is the primary signal used for diagnosing sleep apnea, we use it to explore the underlying patterns and phenotypes in our dataset. In particular, we use fuzzy clustering on two types of features:
\begin{enumerate}
    \item Partial autocorrelation functions of airflow signals
    \item Spectral density functions of airflow signals using Welch's method
\end{enumerate}

These are considered in \cref{sec:pacf}, and \cref{sec:welch} respectively. We note that these methods serve the advantage of converting our airflow signals of varying lengths into features of equal lengths, allowing us to use Euclidean distance for our analysis. We selected the number of clusters, $K$, for each case by considering values in $\{2,3,..,6\}$ and selecting the one giving the highest Silhouette coefficient - the average of the silhouette score over all the data points. In both cases, we ended up with 4 clusters. 

Recall that fuzzy clustering gives a membership probability vector $\{u_{c}\}_{c=1}^K$ for each instance in the dataset. To identify variables that predict these cluster memberships, we use Dirichlet Regression with the 30 variables described in \cref{sec:data} as our covariates. Covariates that are significant with $p$-values less than 0.1 are then explored to identify the patterns or phenotypes in each cluster. 

% We use the alternate parametrization of  Dirichlet distribution in our work as explained in \cref{sec:dirich}

\subsection{Clustering using PACF}
\label{sec:pacf}
To explore the correlations inherent in the airflow timeseries, we considered the autocorrelation function (ACF) and partial autocorrelation function (PACF). Given a time series $\vc{x}=\{x_i\}_{i=1}^T$, we denote its empirical ACF and PACF at lag $t$ by $\rho_{\vc{x}}(t)$ and $\phi_{\vc{x}}(t)$ respectively. Further, let us denote by $\rho_{\vc{x},r}=[\rho_{x}(1),\rho_{x}(2),...,\rho_\vc{x}(r)]$, the ACF vector of the time series up to lag $r$.
Galeano and Pe\~{n}a \cite{acf} used this to define a distance between two time series $\vc{x}$, $\vc{y}$ by setting
$$d_{ACF}(\vc{x},\vc{y})= \sqrt{(\rho_{\vc{x},r}-\rho_{\vc{y},r})^T \ \Omega  \ (\rho_{\vc{x},r}-\rho_{\vc{y},r})}$$
where $\Omega$ is a matrix of weights set by the user, and both time series are assumed to have negligible autocorrelation beyond the $r$th lag. Caiado et al \cite{pacf} then extended this definition to the PACF, similarly defining $$d_{PACF}(\vc{x},\vc{y})= \sqrt{(\phi_{\vc{x},r}-\phi_{\vc{y},r})^T \ \Omega \ (\phi_{\vc{x},r}-\phi_{\vc{y},r})}$$ where $\phi_{\vc{x},r}=[\phi_\vc{x}(1),...,\phi_\vc{x}(r)]$. For our work, we set $\Omega =\mathbb{I}_r$ (i.e the Euclidean metric on the ACF/PACF vectors) and refer to these distances as the ACF and PACF distances of lag $r$ respectively.

\begin{table}[h]
\label{tab:results_pacf}
	\centering
	\small
	\renewcommand{\arraystretch}{1.2}
	\begin{tabular}{|c|c|c|c|c|}
		\hline
		{\bf Cluster} & Cluster 1 & Cluster 2 & Cluster 3 &Cluster 4\\ \hline 
		{\bf Number of patients} & 14  & 3 & 35 & 27 \\ \hline
		{\bf Median \emph{ahi}} & 5.05 & 4.25  &	3.70 & 2.60   \\ \hline 
		{\bf (min \emph{ahi}, max \emph{ahi})}&  (0.6, 14.8)	& (0.8, 113.9) &	(3.5, 4) & (0.6, 40.2)\\
		\hline
		{\bf Percentage with $\emph{ahi}\geq 2$} & 78.6\% &	76.7\% &	100 \% &  63.0\%\\ \hline
	\end{tabular}
	\caption{Results of fuzzy clustering using PACF distance of lag 1500 on airflow signals.  The \emph{ahi} row indicates the percentage of of subjects with $\emph{ahi} \geq 2$ in each cluster.} 
	\label{tab:fuzzypacf}
\end{table}

In our analysis, the ACF for our airflow signals continued to be significant beyond lags as high as $t=60000$, corresponding to a time of 30 minutes. On the other hand, the PACF for all signals are approximately 0 (non-significant) beyond $t=1500$ or a time of about 47s. The reason for this is that the ACF can carry effects of the previous lags indefinitely, while the PACF of a timeseries $\vc{x}$ at time lag $t$ evaluates the relationship between $x_{i}$ and $x_{i+t}$ with the effects of the previous lags removed. For our fuzzy clustering analysis we therefore chose to use the PACF distance of lag $r=1500$. \Cref{tab:results_pacf} summarizes the statistics of the patients' $ahi$ in each of these clusters.
\begin{figure}
	\centering
	\includegraphics[scale=0.7]{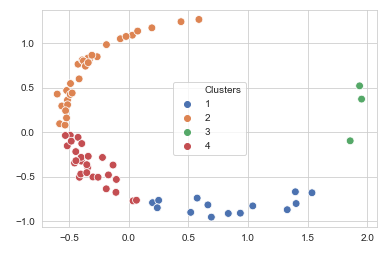}
	\caption{MDS plot obtained using PACF distance (of lag 1500) between airflow signals of OSA patients. The clusters were obtained using fuzzy clustering with 4 clusters and a fuzzy parameter value of $m=1.5$.}
	\label{fig:mds_pacf}
\end{figure}

\begin{figure}
	\centering
	\includegraphics[scale=0.45]{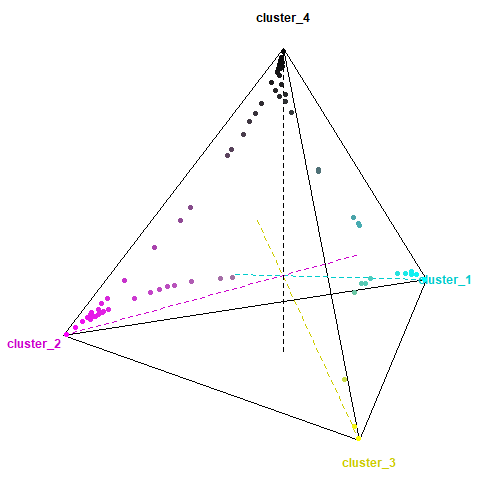}
	\caption{Quaternary plot representing membership of patients to one of four fuzzy clusters obtained using PACF distance (of lag 1500) on airflow signals. The clusters appear fairly well separated as most points are concentrated at the vertices. Note that the three points in cluster 3 may be viewed as outliers and that the other clusters have very small probability components for this cluster.}
\label{fig:quaternary_pacf}
\end{figure}

From the corresponding MDS plot displayed in \cref{fig:mds_pacf}, we note that the three points forming cluster 3 appear to be outliers. This is further supported by quaternary plot, \cref{fig:quaternary_pacf}, which highlights the fact that data points belonging to the other three clusters have very low probability components for cluster 3 (with over 90\% of them being less than 0.1).  A Dirichlet regression analysis on the resulting cluster probabilities shows significant $p$-values for several variables whose mean and standard deviation for each cluster are presented in \cref{tab:variables-pacf}.

\begin{table}[h]
	\small
	\centering
	\renewcommand{\arraystretch}{1.2}\resizebox{\textwidth}{!}{
	\begin{tabular}{|c|c|c|c|c|}
		\hline
		{\bf Predictor variables} & {\bf Cluster 1} & {\bf Cluster 2} & {\bf Cluster 3} &   {\bf Cluster 4}\\ \hline 
		osa-daytime	&	9.64	$\pm$	3.99	&{12.47}	$\pm$	4.50* 	&	8.67	$\pm$	2.08 & 11.37 $\pm$ 3.92*	\\\hline		
		osa-sleep-dist	&	14.00	$\pm$	5.42	&{15.17}	$\pm$	5.80*	&	13.00	$\pm$	4.36 & 12.56 $\pm$ 6.89	\\\hline
		osa-phys-suff	&	13.71	$\pm$	4.46	&	12.80	$\pm$	6.44*	&	15.67	$\pm$	6.66$^\dagger$ & 11.52 $\pm$ 6.08	\\\hline
		osa-caregiver	&	10.42	$\pm$	5.73	&	{14.83}	$\pm$	6.70	&	18.33	$\pm$	4.73 & 12.15 $\pm$ 6.07*\\\hline
		pedsql-phys-func	&	15.36	$\pm$	5.33	&	{17.43}	$\pm$	6.37	&	16.67	$\pm$	6.43 & 21.89 $\pm$ 8.29*	\\\hline
		exercise	&	3.71	$\pm$	1.73	&	{4.13}	$\pm$	1.11	&	5.00	$\pm$	1.73* &  3.89 $\pm$ 1.15**	\\\hline 
	\end{tabular}}
	\caption{Fuzzy clustering on PACF features of airflow: the chart displays the mean $\pm$ SD of the most  significant variables predictors of the fuzzy clusters obtained from a dirichlet regression analysis on 30 covariates. Cluster 1 was the base class for the regression. Symbols $^\dagger$, $*$, $**$  indicate that the variable was significant for the respective cluster with p-values $<0.1$, $<0.05$, $<0.01$ respectively. The p-value of the precision term was close to 0. }
	\label{tab:variables-pacf}
\end{table}

The three patients in the outlier class do not show a consistent pattern in the variables presented. For example, one patient presents a severe case with physical suffering, physical functioning issues, caregiver concern and night time symptom scores far above the median of the whole dataset. Another patient however, shows values for these variables that fall well below the median. It is likely that these patients are outliers due to the characteristics of their airflow signals.

In addition to having the highest median $ahi$, cluster 1 shows the highest physical suffering score which describes symptoms such as: mouth breathing because of nasal obstruction, frequent colds, nasal discharge and difficulty in swallowing foods. In contrast, this cluster also reports the lowest caregiver concern and lowest issues with physical functioning, where questions comprising the latter category inquired about problems with walking, running, participating in sports and energy levels. Additionally, daytime sleepiness symptoms were lowest for this cluster, and nighttime sleep disturbance scores were high, indicating that the primary issues are related to nighttime sleep and breathing.

Cluster 2 shows the most severe issues with nighttime sleep disturbances and daytime sleepiness, along with the highest caregiver concern. This cluster appears to have all the typical symptoms of OSA. Further, this cluster suffers from moderate issues with physical functioning and physical suffering. Overall, it appears to be the most severe of clusters 1, 2 and 4, and may be considered a priority for being referred to a sleep specialist.

Lastly, cluster 4 seems to be characterized by severe physical functioning issues but a low median ahi. This cluster suffers from moderate daytime sleepiness symptoms and has the lowest physical suffering and nighttime sleep disturbance scores. Due to atypical symptoms that primarily affect physical functioning, patients in this cluster may benefit from seeing a physical therapist.

% to define a distance between time series, where r is fixed to be a time lag such that \rho_\vc{x}(t)\cong 0 for all t>r. In particular, they set the distance two time series \vc{x}, \vc{y}
%d_{ACF}(\vc{x},\vc{y})= \sqrt{(\rho_{\vc(x),r}-\rho_{\vc(x),r})^T \Omega (\rho_{\vc(x),r}-\rho_{\vc(x),r})}
%where \vc{x}, \vc{y} are two time series with negligible autocorrelation beyond the r-th lag. Similarly, Caiado extended this definition to the PACF defining
%d_{PACF}(\vc{x},\vc{y})= \sqrt{(\psi_{\vc(x),r}-\psi_{\vc(x),r})^T \Omega (\psi_{\vc(x),r}-\psi_{\vc(x),r})}
%
%Galeano and Peno considered the use of the ACF vector of 

\subsection{Clustering using spectral density}
\label{sec:welch}

Besides exploring airflow signals in the time domain, we also considered frequency-domain features, specifically,  spectral density obtained using Welch's method \cite{welch}. This method breaks up a signal into successive (possibly overlapping) blocks of a fixed size, evaluates the periodogram or spectral density for each of these blocks and then averages them using a chosen window function. There are several advantages to using this over the periodogram: 1) one can control the length of the spectral density output, useful for very long time series, 2) averaging over multiple blocks reduces variance in the spectral density estimate, 3) the spectral density output of all time series (of varying lengths) is with respect to a common set of frequencies. 

\begin{table}[h]
	\centering
	\small
	\renewcommand{\arraystretch}{1.2}
	\begin{tabular}{|c|c|c|c|c|}
		\hline
		{\bf Cluster} & Cluster 1 & Cluster 2 & Cluster 3 &Cluster 4\\ \hline 
		{\bf Number of patients} & 10  & 13 & 34 & 17 \\ \hline
		{\bf Median \emph{ahi}} & 14 & 3.8  &	3.8 & 4   \\ \hline 
		{\bf (min \emph{ahi}, max \emph{ahi})}&  (1.6, 40.2)	& (1.6, 10) &	(0.6, 113.9) & (0.6, 15.2)\\
		\hline
		{\bf Percentage with $\emph{ahi}\geq 2$} & 90\% &	76.9\% &	73.5 \% &  58.8\%\\ \hline
	\end{tabular}
	\caption{Results of fuzzy clustering using spectral density of airflow signals with Welch's method (Tukey window of 128s).  The \emph{ahi} row indicates the percentage of of subjects with $\emph{ahi} \geq 2$ in each cluster.} 
		\label{tab:results_welch}
\end{table}

\begin{figure}[h]
	\centering
	\includegraphics[scale=0.7]{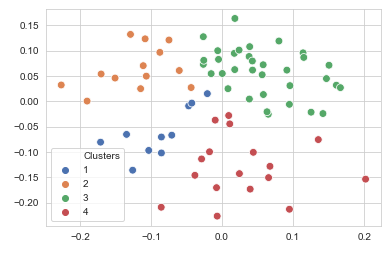}
	\caption{MDS plot obtained using Euclidean distance on (Welch's) spectral density estimates on airflow signals of OSA patients. The clusters were obtained using fuzzy clustering with 4 clusters and a fuzzy parameter value of $m=1.5$.}
	\label{fig:mds_welch}
\end{figure}

\begin{figure}[h]
	\centering
	\includegraphics[scale=0.45]{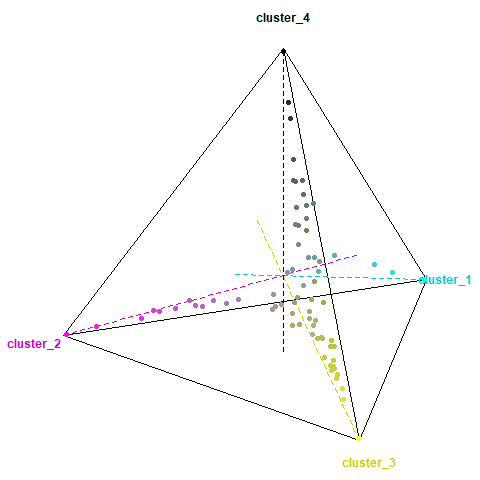}
	\caption{Quaternary plot representing membership to each of four fuzzy clusters obtained using Euclidean distance on (Welch's) spectral density estimates on airflow signals We note that this case is not as well-separated as the one presented in \cref{fig:quaternary_pacf}. }
	\label{fig:quaternary_welch}
\end{figure}

In our work, we used a Tukey window of 128s or $2^{12}$ points for our airflow signal sampled at a frequency of $2^5$ Hz. This gave an output spectral power vector of length 2049 which we normalized to obtain the spectral density. We then applied fuzzy clustering with Euclidean distance to obtain the final clusters. We summarize the statistics of the $ahi$ scores in \cref{tab:results_welch}.
The corresponding MDS and quaternary plots are displayed in \cref{fig:mds_welch} and \cref{fig:quaternary_welch} respectively. 

%\begin{figure}
%	\centering
%	\includegraphics[scale=0.5]{ImagesII/mds_welch.png}
%	\caption{MDS plot of airflow (sleep cycle 2) with Dynamic Time Warping distance. The clusters are obtained from fuzzy 3-clustering.}
%	\label{fig:mds_pacf}
%\end{figure}

\Cref{tab:variables-welch} presents the most significant variables predicting the fuzzy clustering components based on a Dirichlet regression analysis, along with the means and standard deviations for each cluster. We note that all four clusters report fairly high scores on night time sleep disturbance symptoms - a classic symptom of OSA.

\begin{table}[h]
	\small
	\centering
	\renewcommand{\arraystretch}{1.2}\resizebox{\textwidth}{!}{
	\begin{tabular}{|c|c|c|c|c|}
		\hline
		{\bf Predictor variables} & {\bf Cluster 1} & {\bf Cluster 2} & {\bf Cluster 3} &   {\bf Cluster 4}\\ \hline 
		cshq-sleep-hrs	&	9.55	$\pm$	2.30	&{9.58}	$\pm$	1.97$^\dagger$ 	&	9.71 $\pm$	1.59$^\dagger$   & 10.03 $\pm$ 1.71	\\\hline		
		cshq-night-waking	&	12.08	$\pm$	9.71	& 13.96	$\pm$	18.75 	&	13.78	$\pm$ 14.13$^\dagger$ & 10.89 $\pm$ 8.06	\\\hline		
		cshq-night-sleep	&	76.5	$\pm$	12.62	& 66.85	$\pm$	10.13 	&	68.88	$\pm$ 9.90$^\dagger$ &  62.12 $\pm$ 8.77**\\\hline		
		cshq-day-sleep	&	14.20	$\pm$	1.81	& 15.23	$\pm$	1.92 	&	15.00	$\pm$ 1.18$^\dagger$  & 14.76 $\pm$ 1.48	\\\hline		
		osa-emotional	&	11.9 $\pm$	5.97	&	{8.38} $\pm$	4.52*	&	12.21	$\pm$	5.84$^\dagger$ & 8.88 $\pm$ 5.58 \\ \hline
		osa-overall	&	4.9	$\pm$	1.73	&	3.85	$\pm$	2.08	&	3.76	$\pm$	1.69$^\dagger$ & 2.82 $\pm$ 1.51\\\hline
		pedsql-phys-func	& 22.80	$\pm$	5.85	&	22$\pm$	9.07	&	17.15	$\pm$	5.95** & 16.59 $\pm$ 7.81**	\\\hline
		pedsql-social-func	&	13.80	$\pm$	3.74	&	11.77	$\pm$	5.05$^\dagger$	&	11.85	$\pm$	4.18 & 9.47 $\pm$ 4.46	\\\hline
		pedsql-school-func	&	15.80	$\pm$	5.43	&	12.62	$\pm$	5.17	&	13.45	$\pm$  4.09 & 13.94 $\pm$ 5.04***	\\\hline
		wake	&	22.47	$\pm$	15.93	&	{18.16}	$\pm$	8.13	&	14.69	$\pm$	8.31*  & 21.21 $\pm$ 15.01	\\\hline
		NREM2	&	34.44	$\pm$	13.92	&	37.05	$\pm$	10.14	&	35.98 $\pm$	9.60$^\dagger$ & 34.56 $\pm$ 11.48$^\dagger$	\\\hline	
		sleep-efficiency	&	77.62	$\pm$	15.96	&	81.89	$\pm$	8.18	&	85.59	$\pm$	7.97$^\dagger$ &  78.87 $\pm$ 15.02	\\\hline 
	\end{tabular}}
	\caption{Fuzzy clustering on spectral density: the chart displays the mean $\pm$ SD of the most  significant variables predictors of the fuzzy clusters obtained from a dirichlet regression analysis on 30 covariates. Cluster 1 was the base class for the regression. Symbols $^\dagger$, $*$, $**$, $***$  indicate that the variable was significant for the respective cluster with p-values $<0.1$, $<0.05$, $<0.01$, $<0.001$ respectively. The p-value of the precision term was $<0.001$. }
\label{tab:variables-welch}
\end{table}

Cluster 1 shows the most severe case of all clusters, with the highest median $ahi$ by far and 90\% of patients exhibiting $ahi\geq 2$. This group has the highest nighttime symptom score based on the CSHQ survey, high levels of emotional stress and the worst issues with physical, social and school functioning. Further, this group also suffers from low sleep efficiency and a high percentage of wake sleep stages. Due to the severity of symptoms, patients in this cluster may be considered as a priority in being referred to the sleep specialist. Additionally, due to the diversity of symptoms experienced, these patients may need a multi-factorial treatment involving other diagnoses. 

In contrast, cluster 4 shows a high number of sleep hours and low scores on issues with physical, social and school functioning. However, this cluster also reports a low average quality of life rating and a high percentage of wake stages. Out of all four clusters, this one has the lowest percentage of patients with $ahi\geq 2$ and may be considered the mildest case. It appears that unlike cluster 1, these patients primarily suffer from sleep issues which do not affect the other aspects of their life to the same extent. 

Clusters 2 and 3 show results that are intermediate to Clusters 1 and 4, both reporting high sleep efficiencies. Cluster 3 stands out with a high response to emotional stress survey questions, despite having a low percentage of wake sleep stages. Children in this cluster may benefit from being referred to a psychiatrist in order to study the connection between their emotional symptoms and their sleep quality. 
Patients in cluster 2 on the other hand have relatively low emotional and school functioning issue scores,but report high physical suffering scores. These patients may benefit from a dentist referral - for example, there are multiple patients with very high CFI in this groups. Overall, patients in these two clusters may need specific treatments to address their individual issues. 

\section{TDA analysis of airflow signals }
\label{sec:tda_analysis}

In this section, we present an analysis of OSA patients based on the topological features of their airflow signals. To conduct TDA on these signals, we first transformed them into periodicity scores over time frames i.e. we divided each time series into blocks or frames of fixed time lengths and calculated the periodicity for each block. This was done using two steps: (1) transforming time series into point cloud data using a sliding window and (2) calculating periodicity over time frames of the point cloud data through 1D persistence.

We followed this up with a three-fold analysis of the periodicity score data by: 1) clustering using zero dimensional persistence (complete-linkage hierarchical clustering), 2) applying ANOVA to evaluate periodicity sores over time and clusters, 3) clinical phenotyping of obtained clusters using linear discriminant analysis~(LDA). The ANOVA and LDA were carried out using IBM SPSS Statistics 26 for Windows, NY: IBM Corp.

To accommodate our airflow signals are of varying lengths (between 2 to 10 hours sampled at 32Hz), we reinterpolated the time series so they have the same number of time samples ($1.2\times 10^6$ samples). As these signals were too long to apply TDA, we divided them into frames of 10 minutes each, giving a total of 62 time frames. Due to this producing a high number of features relative to our 74 samples, we only chose to analyse the 31 odd frames. Using this we created the trajectory matrix in \cref{eqn:trajec} with $n=1200$ and $p=14$. This number of points $n$ was chosen by trial and error, so that the number is large enough for loops to appear, but is feasible in terms of computational time.  The delay parameter $\tau$ was calculated for each subject by each frame. As autocorrelation of each frame of each patient was larger than $2/\sqrt{T}$, we selected the time lag whose autocorrelation is  minimum, but not zero. The time lags with minimum autocorrelation in  all 31 frames of all 74 patients  ranged from 18 to 42. 

\begin{figure}[h]
\centering	
\includegraphics[scale=0.21]{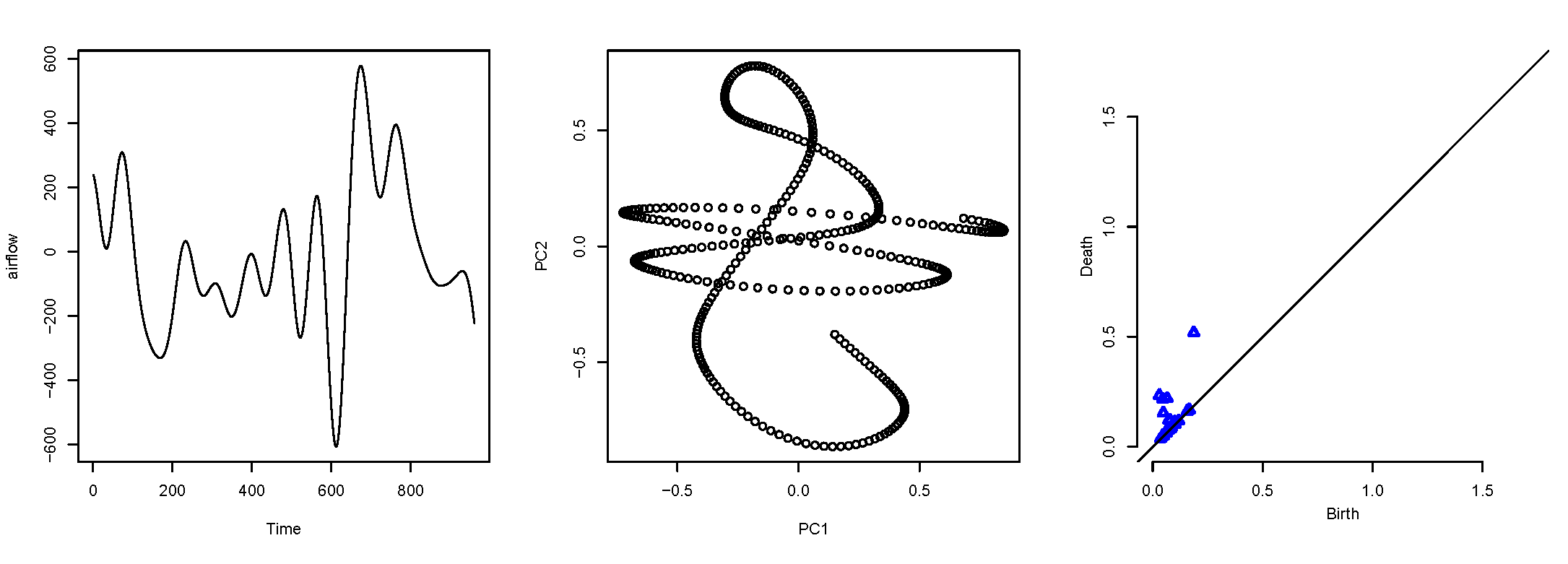}  \includegraphics[scale=0.21]{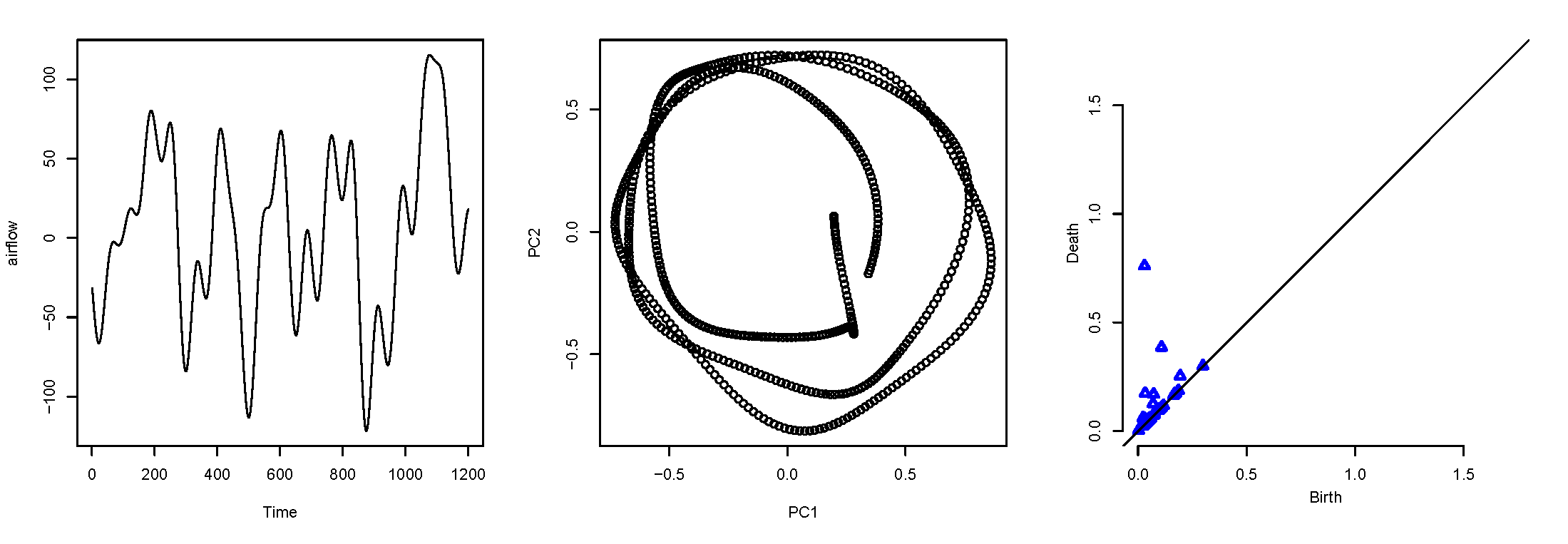}\\
\includegraphics[scale=0.21]{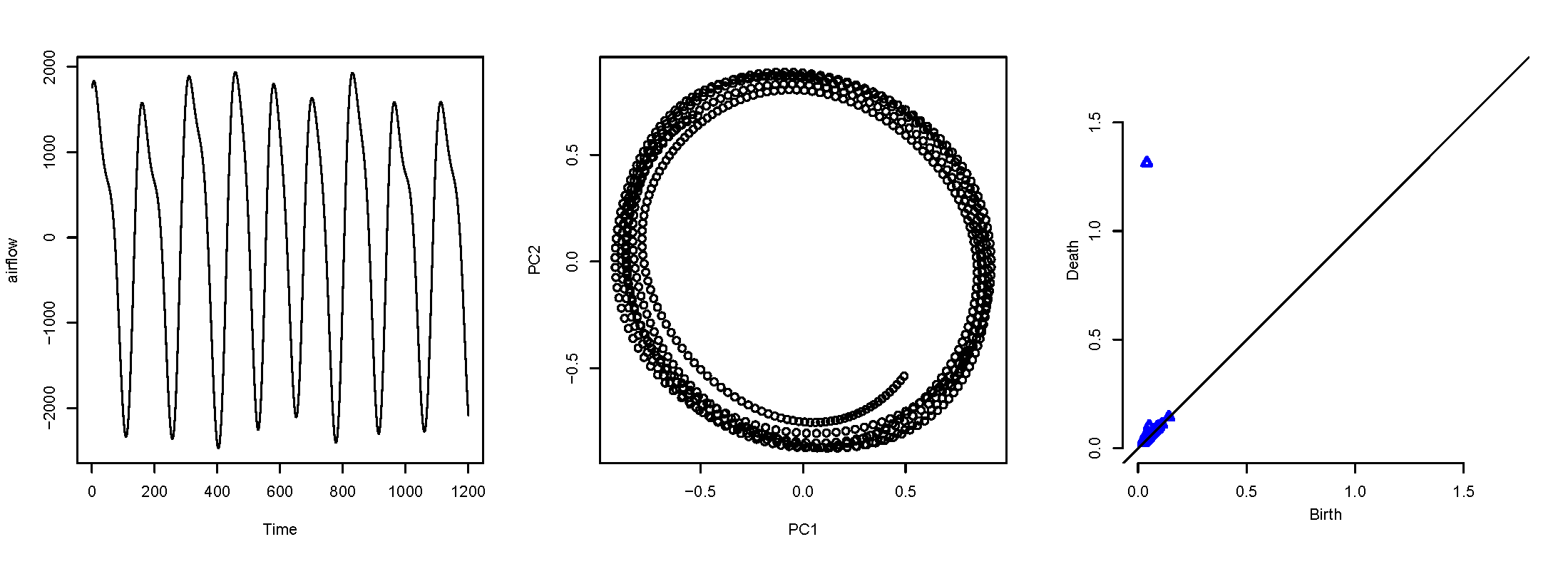}
\includegraphics[scale=0.21]{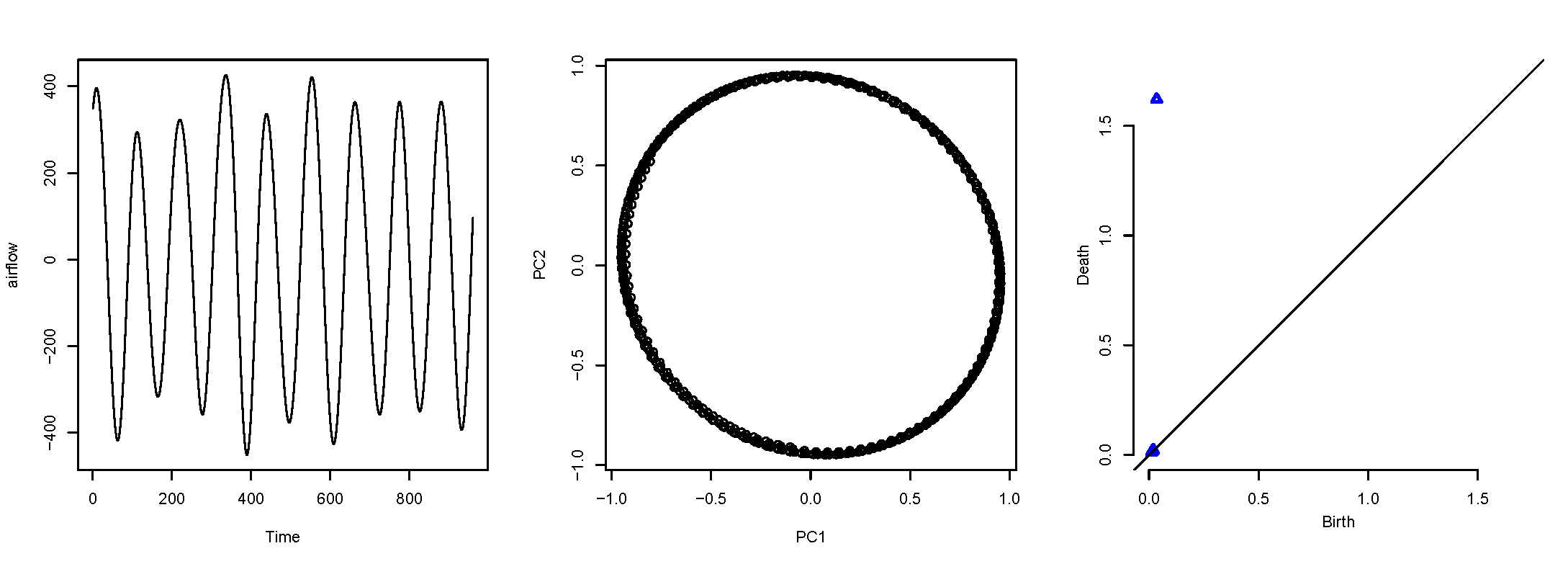}
\caption{Examples of TDA on time series: extracting a single frame of the time series, scatter plot of first two principal components, and the corresponding persistence diagram. (From top to bottom and left to right): Periodicity scores are 0.19, 0.42, 0.74 and 0.92.}
\label{fig:exTypical}
 \end{figure}

The authors in \cite{perea15B} suggested using moving average to denoise the timeseries as an option, however, as our re-interpolated airflow signals were fairly smooth, we chose to omit this step. However, we did standardize each point cloud as suggested by the authors in \cite{perea15B}. 
We present visual representations of the process in \cref{fig:exTypical}, by providing examples of a single time frame and the corresponding 2D-point cloud and persistence diagram for 4 patients.

As the periodicity scores over the 31 time frames form a discretized Morse function, we applied super-level set filtration as described in \cref{sec:ph}. We then calculated pairwise distances between their corresponding zero-dimensional persistence diagrams using the bottleneck distance defined in \cref{eqn:bottleneck}. This gives us a hierarchical clustering (see \cite{PH0hierarchical} for comparison between hierarchical clustering and 0-dimensional persistent homology)  and we obtained four clusters at distance height $0.3$ in the dendrogram. In \cref{tab:cluster-PH0}, we report summary statistics of periodicity score and \emph{ahi}. We first calculate average periodicity score and standard deviation of 31 frames for each subject, thereby obtaining 74 averages and their corresponding 74 standard deviations. We then found the mean and standard deviation of these 74 averages - calling this the ``Avg. period score $\pm$ SD", presented in row 3 of \cref{tab:cluster-PH0}. Additionally, the 74 standard deviations we found (of periodicity scores over the 31 frames for each patient), were averaged within each cluster to give the fourth row of \cref{tab:cluster-PH0}, denoted as ``Avg. frame-SD".

The clustering shows interesting patterns. Cluster 4 consists of patients with the highest average periodicity scores while being lowest in both frame-SD and SD of \emph{ahi}.  Overall we find that higher frame-SDs are associated with lower periodicity scores and higher \emph{ahi} scores. This finding is not  surprising since apnea and hypopnea events are derived from the airflow. 

Based on our findings, periodicity of airflow may be used as a basis for future studies on using breathing patterns to screen pediatric OSA patients. For example, unlike PSG which is hard to access and expensive, breathing cycle data may be conveniently measured using smart wearable devices \cite{wearable}. 
\begin{table}[h!]
	\small
	\centering
	\renewcommand{\arraystretch}{1.2}\resizebox{\textwidth}{!}{
	\begin{tabular}{|c|c|c|c|c|}
		\hline
		{\bf Cluster}& 1 &  2 & 3 & 4\\ \hline  
{\bf Number of subjects} & 30~(40\%) & 14~(19\%) & 13~(18\%) &17~(23\%) \\ \hline	{\bf Avg. Period score $\pm$ SD} & 0.59 $\pm$ 0.12 & 0.59 $\pm$ 0.05 & 0.53 $\pm$ 0.05 & 0.65 $\pm$ 0.12 \\ \hline
{\bf Avg. frame-SD $\pm$ SD} & 0.22 $\pm$ 0.03 & 0.24 $\pm$ 0.03 &0.25 $\pm$ 0.02 &0.18 $\pm$ 0.03 \\ \hline
		{\bf average \emph{ahi} $\pm$ SD } & 5.98 $\pm$ 7.04 &8.89 $\pm$ 9.85  & 14.81 $\pm$ 30.16 & 4.49 $\pm$ 4.28 \\ \hline	
	\end{tabular}}
	\caption{Summary statistics for clusters obtained from zero-dimensional PH.  The average and SD in the third row are taken after averaging the scores of 31 frames per patient. The fourth row presents average of standard deviations taken over 31 frames. Lower the average frame- SD, higher average periodicity score and lower average \emph{ahi}. }
	\label{tab:cluster-PH0}
\end{table}

\begin{figure}[h!]
\centering	
 \includegraphics[scale=0.55]{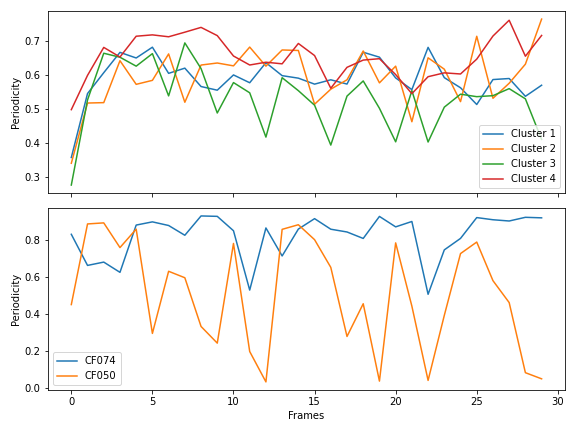} 
% \includegraphics[scale=0.20]{ImagesII/profilePH0.pdf} 
% &\includegraphics[scale=0.20]{ImagesII/DemoCF074CF050B31.pdf}
\caption{(Top) Profile plot of average periodicity over frames in each cluster.The clusters are determined from Dim-0 PH filtration of periodicity scores of time frame. (Bottom) Profile plot of  two patients (CF074 and CF050). The patient CF074's \emph{ahi} 2.9 and The patient CF050's \emph{ahi} 113.9, highest index among all 74 patients.}
\label{fig:profilePH0}
 \end{figure}

For better understanding of the patterns of periodicity scores over time frames and clusters simultaneously, we carry out repeated measures analysis of variance (ANOVA), where the time frame is a within-subject factor and cluster a between-subject factor. Anova shows that each predictor variable, cluster (p-value$<0.001$), and time frame (p-value $=0.012$) are statistically significant. The interaction of these two variables (p-value$=0.125$) shows weak evidence that the periodicity scores within clusters are not independent over time frames. The pattern of interaction between cluster and time frame on the periodicity score can be visualised in the profile plot on the top in Figure \ref{fig:profilePH0}. For example, in cluster 4, the magnitude of oscillation is smaller than other clusters  while in cluster 3, the magnitude of oscillation is higher than others overall over time frames, and the pattern of oscillation is similar from frame to frame, especially within frames 5 and 22. 

We pick two patients, CF074, who has a low \emph{ahi} of 2.9 and CF050, who has the highest \emph{ahi} of $113.9$, and display their periodicity scores over time frames on the bottom in Figure \ref{fig:profilePH0}.  The periodicity score of  patient CF050 has higher fluctuations compared to patient CF074. Additionally, the average periodicity score for CF050 is lower than that for CF074. One may also visualise this pattern by observing the way the 1D homology (loops) of these patients changes over sleep time.

\begin{figure}[h!]
\centering	
\includegraphics[scale=0.2]{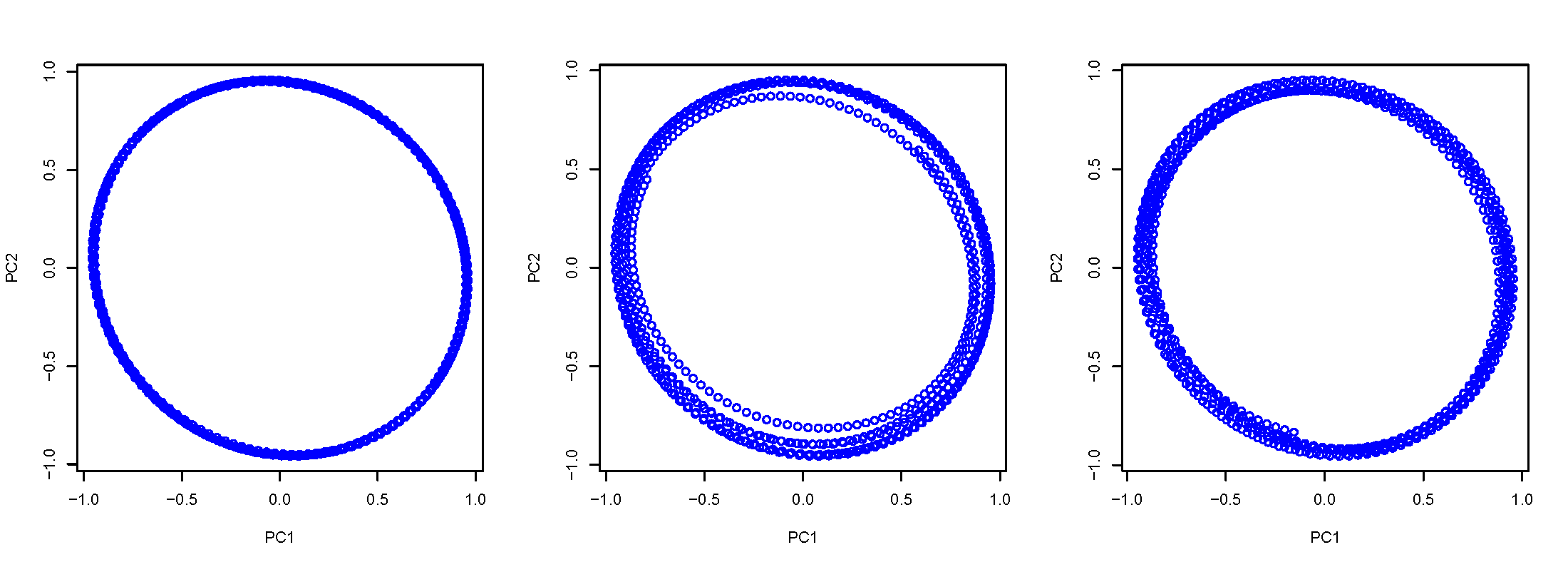} \includegraphics[scale=0.2]{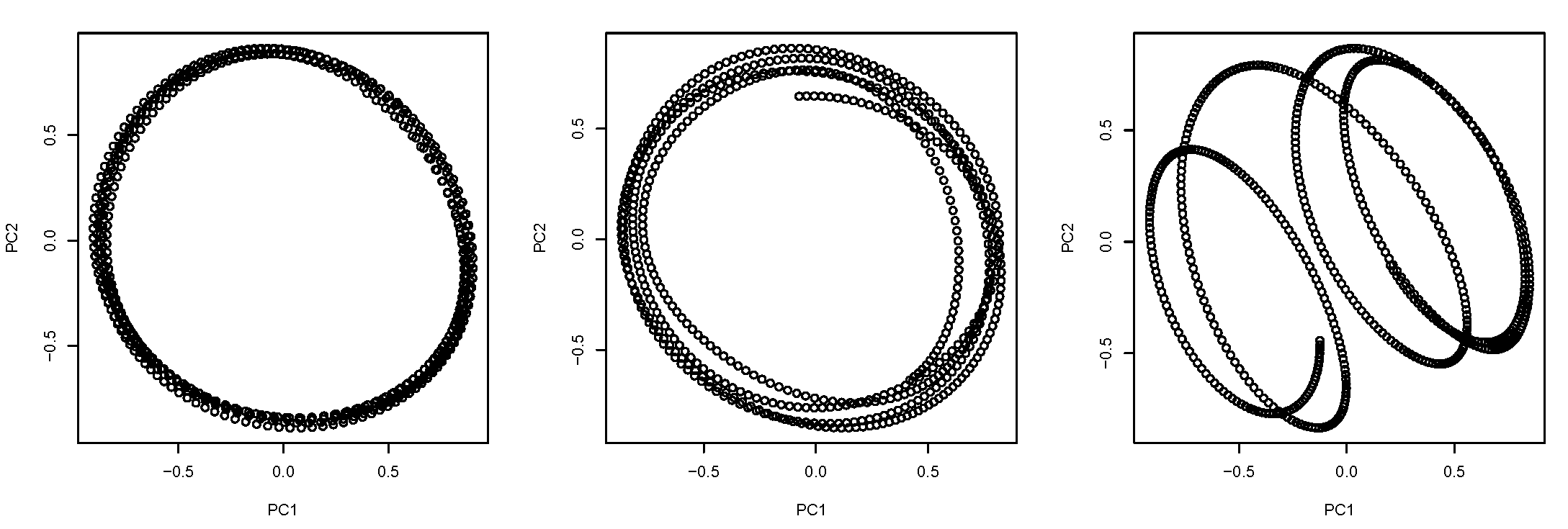}
\includegraphics[scale=0.2]{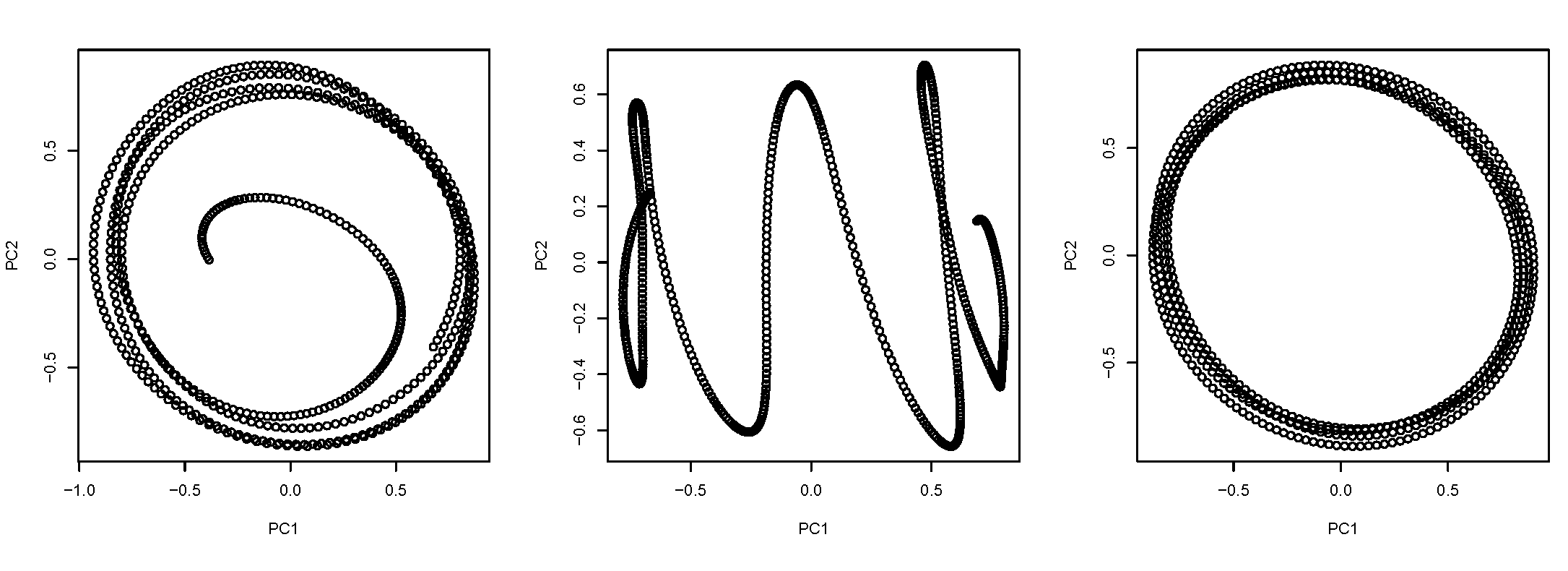}
\includegraphics[scale=0.2]{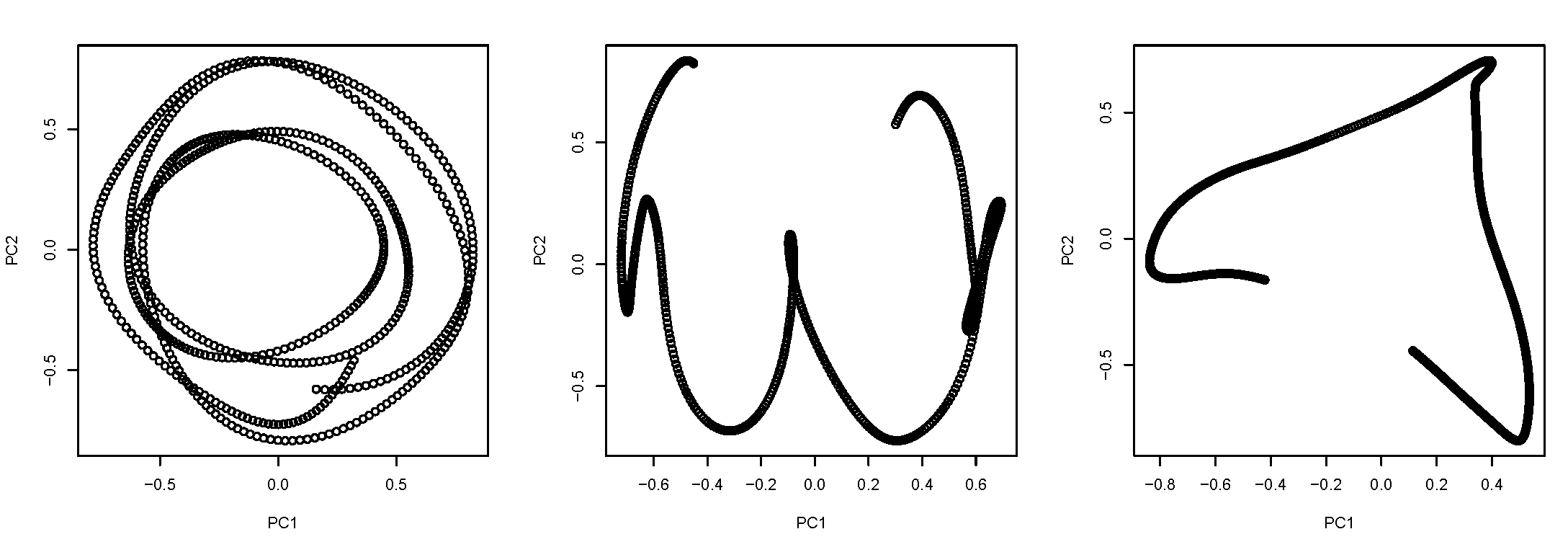}
\caption{Examples of periodicity plots over frame: (Top left-in blue figure) The patient CF074 (male). This pattern and similar to this keep repeating during his sleep showing his breathing is regular and periodic. (Top right and bottom row): The patient CF074 (male). The pattern shows irregular periodicity likely reflects irregularity in breathing.}
\label{fig:variation}
 \end{figure}

\begin{table}[h]
	\small
	\centering
	\renewcommand{\arraystretch}{1.2}\resizebox{\textwidth}{!}{
\begin{tabular}{|c|c|c|c|c|}
\hline
		{\bf Predictors } & {\bf Cluster 1}& {\bf Cluster 2} & {\bf Cluster 3} & {\bf Cluster 4} \\ \hline 
activity (days)	&	4.43	$\pm$	2.03	&	4.29	$\pm$	2.13	&	5.7	$\pm$	2.06	&	5.53	$\pm$	1.94	\\\hline
waso	&	58.08	$\pm$	27.51	&	56.51	$\pm$	32.28	&	89.7	$\pm$	49.01	&	73.01	$\pm$	75.15		\\\hline
total stage-changes	&	70.43	$\pm$	17.60	&	80.36	$\pm$	15.07	&	83.8	$\pm$	32.03	&	67.00	$\pm$	21.15		\\\hline
sleep awakenings	&	21.43	$\pm$	7.29	&	25.64	$\pm$	7.03	&	25.9	$\pm$	10.44	&	19.88	$\pm$	8.05	\\\hline
\end{tabular}}
	\caption{LDA of single-linkage clustering (zero-dimensional PH): the chart shows Mean $\pm$ SD of the most significant predictor variables from a linear discriminant analysis using 30 covariates (see \cref{sec:data}). All the variables have LDA p-values less than 0.1, except waso, which has a p-value of 0.18. The Manova p-value for this table is $0.10$
}
\label{tab:lda-Hclustering-PH-dim0}
\end{table}
For clinical phenotyping, we applied LDA to the clusters (as a dependent variable) using the covariables described in \cref{sec:data}. Four out of a total of 30 predictor variables are show to be effective in classifying 4 clusters based on  p-values $<0.20.$ The averages and SD of each variable and ANOVA p-values are presented in Table \ref{tab:lda-Hclustering-PH-dim0}. Overall, all four predictor variables are moderately significant in distinguishing 4 clusters.
The cluster 4 groups the patients with high activity and lowest in total sleep stage changes and sleep awakenings, yet second highest in waso. The cluster 3 consists the patients with highest values in all four predictor variables. Theses patients are high in daily activities yet have sleep related issues. Theses patients could be prioritised to make a referral to a sleep specialist. Clusters 1 and 2 are less distinguishable. The average values are similar except total sleep changes higher in cluster 2. The patients in cluster 1 might be considered as no risk of OSA.

% \begin{table}
% 	\small
% %\centering
% 	\renewcommand{\arraystretch}{1.0}
% 	%   \begin{tabular}{|p{2.4cm}|p{1.7cm}|p{1.7cm}|p{1.7cm}|p{1.7cm}|p{1cm}|}
% 	\begin{tabular}{|c|c|c|c|c|c|}
	
% \hline
% 		{\bf Predictors } & {\bf Cluster 1~(30)}& {\bf Cluster 2~(14)} & {\bf Cluster 3~(13)} & {\bf Cluster 4 ~(17)} & {\bf p-value$^\dagger$}\\ \hline 
% activity (days)	&	4.43	$\pm$	2.03	&	4.29	$\pm$	2.13	&	5.7	$\pm$	2.06	&	5.53	$\pm$	1.94	&	0.10	\\\hline
% waso	&	58.08	$\pm$	27.51	&	56.51	$\pm$	32.28	&	89.7	$\pm$	49.01	&	73.01	$\pm$	75.15	&	0.18	\\\hline
% sleep totalstage changes	&	70.43	$\pm$	17.60	&	80.36	$\pm$	15.07	&	83.8	$\pm$	32.03	&	67.00	$\pm$	21.15	&	0.09	\\\hline
% sleep awakenings	&	21.43	$\pm$	7.29	&	25.64	$\pm$	7.03	&	25.9	$\pm$	10.44	&	19.88	$\pm$	8.05	&	0.09	\\\hline
% \end{tabular}
% 	\caption{LDA of single-linkage clustering (zero-dimensional PH)  The chart shows Mean $\pm$ SD of the most significant predictor variables from an LDA analysis out of a total of 30 variables. 
% 	Manova p-value for all 4 variables is $0.10.$
% }
% \label{tab:lda-Hclustering-PH-dim0}
% \end{table}

\section{Conclusion and future research }\label{sec:conclusion}

The goal of this work was to conduct clustering analysis on OSA patients, using their airflow signals, in an attempt to explore the possible underlying clinical phenotypes. The features for clustering were obtained in three ways: in the time domain using the PACF up to lag 1500, in the frequency domain using (Welch's) spectral density estimates, and using TDA, specifically, persistent homology. For the former two methods, we used fuzzy clustering with four clusters, and then used the output probability vectors as the dependent variable in a Dirichlet regression analysis using 30 covariates (described in \cref{sec:data}. We then considered the significant variables in the regression (with p-value less than 0.1) to study the patterns in each cluster.

 In the case of PACF, one of the clusters we obtained, namely cluster 3, appears to contain outliers based on \cref{fig:quaternary_pacf} but does not show a consistent pattern across the three patients it contains. The other three clusters may be phenotyped as follows: 
\begin{itemize}\setlength\itemsep{0em}
	\item Cluster 1 is characterized primarily by nighttime sleep and breathing issues, classic symptoms of OSA.
	\item Cluster 2 shows the most severe symptoms overall, and has severe issues with nighttime sleep disturbances and daytime sleepiness and moderate issues with breathing and physical functioning.
	\item Cluster 4 is characterized primarily by severe issues with physical functioning and moderate daytime sleepiness symptoms.
\end{itemize}
Due to the severity and type of symptoms clusters 1 and 2 may be prioritized in a sleep specialist referral. In countries like Canada, the wait time to see a sleep specialist can be up to 2 years. In such a situation, prioritizing patients based on severity may be important in getting them timely treatment. Additionally, cluster 2 may require multi-factorial treatment to address the variety of symptoms they suffer from. Lastly, cluster 4 may benefit from alternative treatments such as physiotherapy.

 Clustering the spectral density obtained from Welch's method, gave the following phenotypes:
\begin{itemize}
\setlength\itemsep{0em}\item Cluster 1 is the most severe of cases with high levels of nighttime sleep disturbances and emotional stress and moderate scores on other symptoms. 
\item  Cluster 2 is characterized primarily by high physical suffering scores, and several show high CFI. 
\item Cluster 3 is characterized by high levels of emotional stress but have high sleep efficiency. 
\item Cluster 4 presents the mildest case out of the four clusters and is primarily characterized by a high percentage of Wake sleep stages and a low mean quality of life. \end{itemize}
It appears that cluster 1 may be prioritized in a sleep specialist referral, and again, due to the diverse nature of their symptoms, may require a multi-factorial treatment. Cluster 2 and 3 may benefit from a referral to a dentist, and a psychiatrist, respectively to study the connection between their OSA and other symptoms.

For clustering using TDA, we divided each airflow signals into time frames of 10 minute each. Each time frame for each patient was then converted into point cloud data by applying sliding window delay embedding. We then calculated the periodicity score of each frame by measuring the size of the most persistent 1-dimensional hole (loop) of the point cloud. These scores were analyzed in two ways: first, we obtained 0-dimensional persistence by performing super level set filtration on the scores over each time frame; we then calculated bottleneck distances between 0-dimensional persistence diagrams, which was used to obtain a hierarchical clustering. 

For the clustering, we chose to cut the dendrogram at a height that gave four clusters and compared periodicity and \emph{ahi} scores across clusters. From our findings, higher severity in OSA is associated with lower periodicity scores, higher variation (average of frame-SD) and higher irregularities in airflow.
 This finding is somewhat expected as the severity of OSA is determined by \emph{ahi} which measured from airflow.   It seems that TDA shows potential in discerning respiratory patterns among OSA patients, and could possibly be used to analyze breathing rates of patients obtained from digital wearable smart devices to assess severity of their symptoms. This would ensure that patients are diagnosed in an inexpensive and timely manner.

\section{Acknowledgement}
%\begin{acknowledgement}
GH would like to thank the National Sciences and Engineering Research Council of Canada (NSERC DG 2016-05167),
Seed grant from Women and Children's Health Research Institute, Biomedical Research Award from American Association of Orthodontists Foundation, and the McIntyre Memorial fund from the School of Dentistry at the University of Alberta.
%\end{acknowledgement}
%\section*{References}
%Here are two sample references: %\cite{Feynman1963118,Dirac1953888}.
\bibliography{mybibfile}
\newpage
\section{Appendix}
This section contains figures referred to in the main article.

\begin{figure}[h!]
\centering
\includegraphics[scale=0.14]{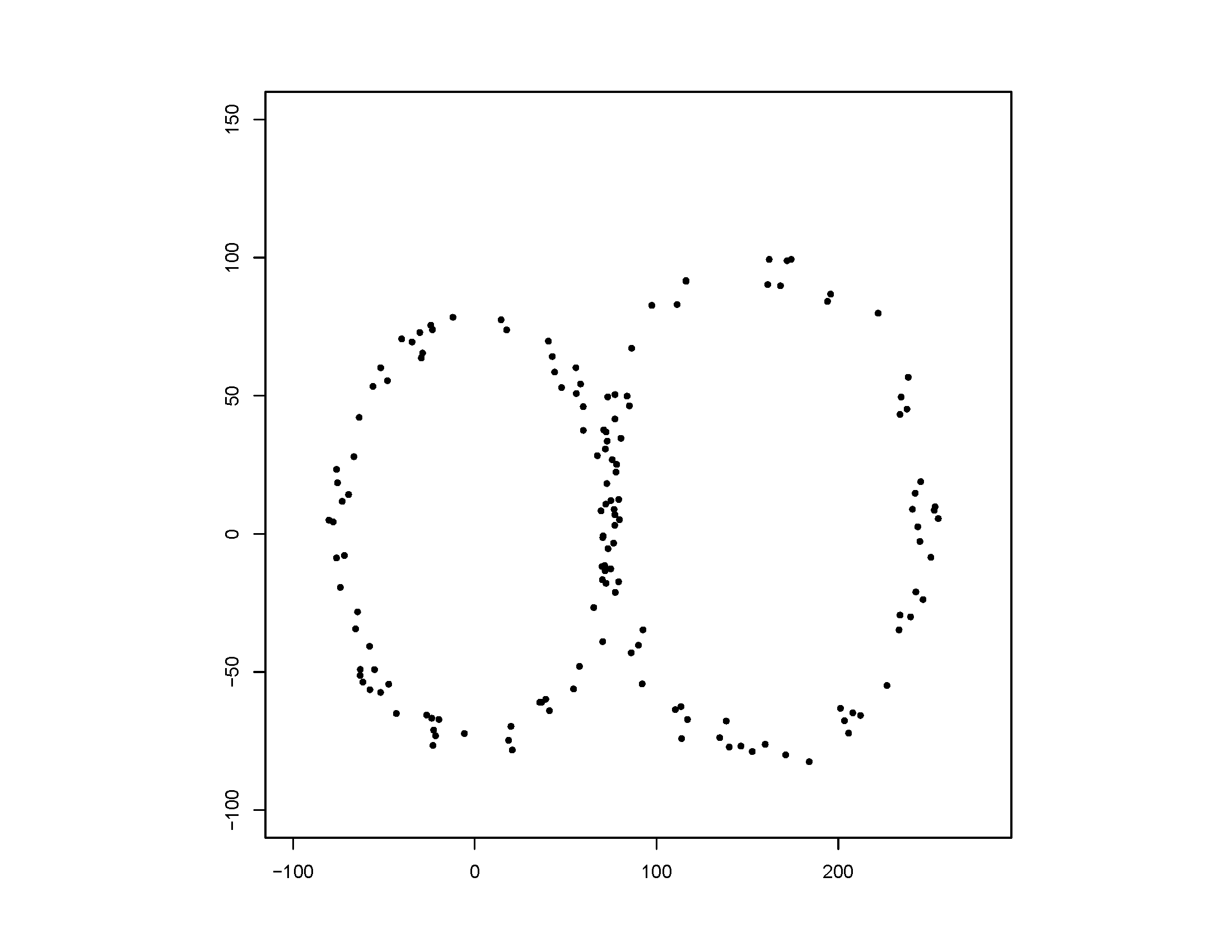} 
\includegraphics[scale=0.14]{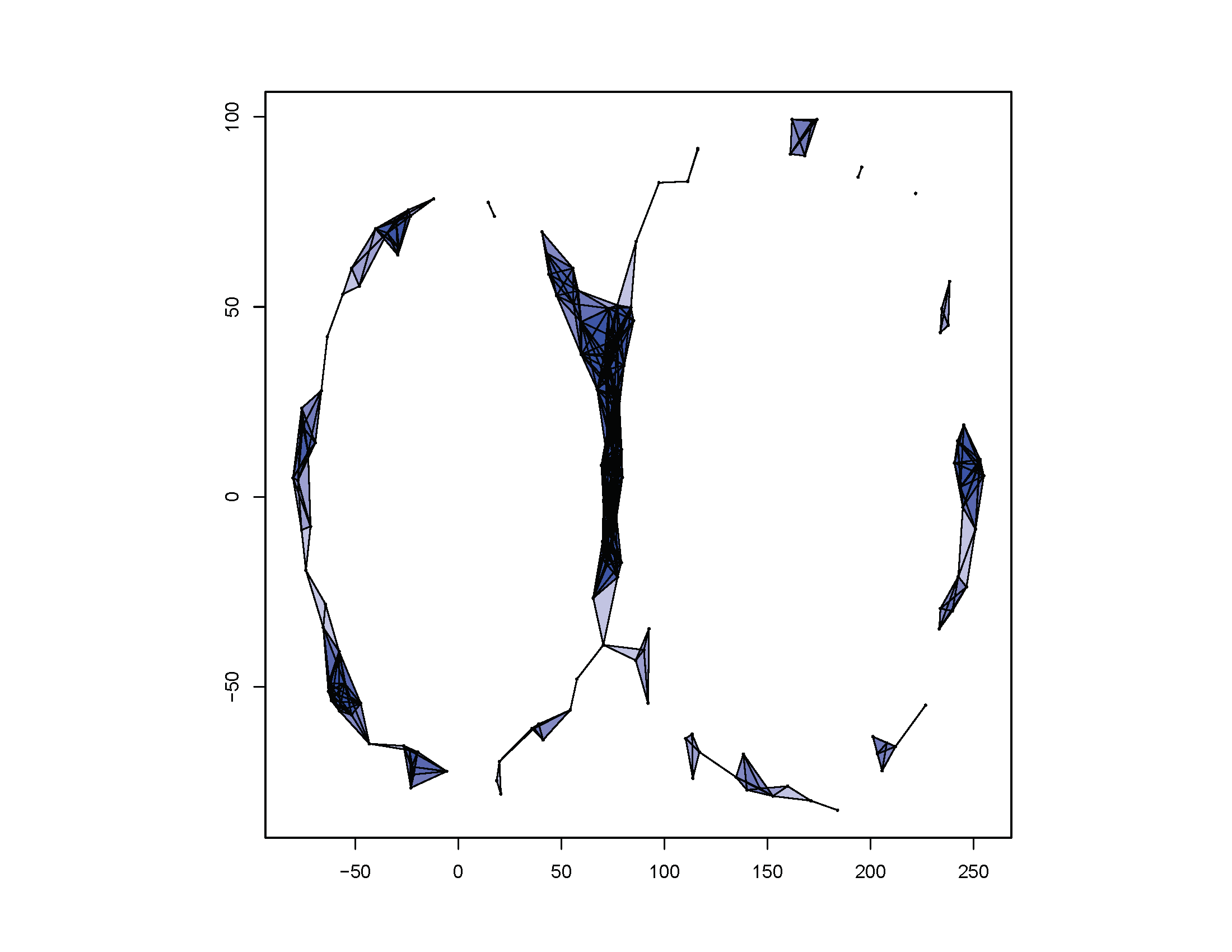}
\includegraphics[scale=0.14]{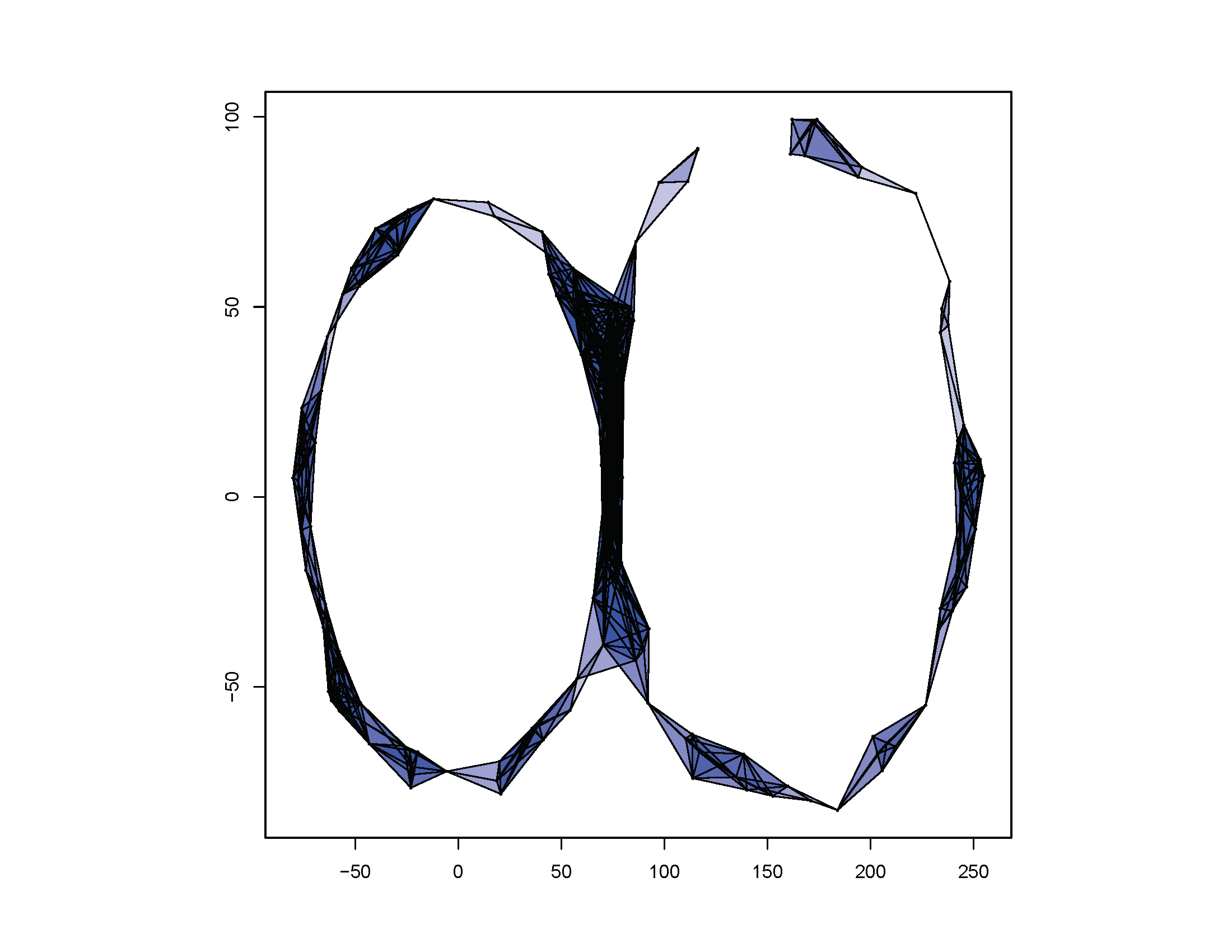}\\
\includegraphics[scale=0.13]{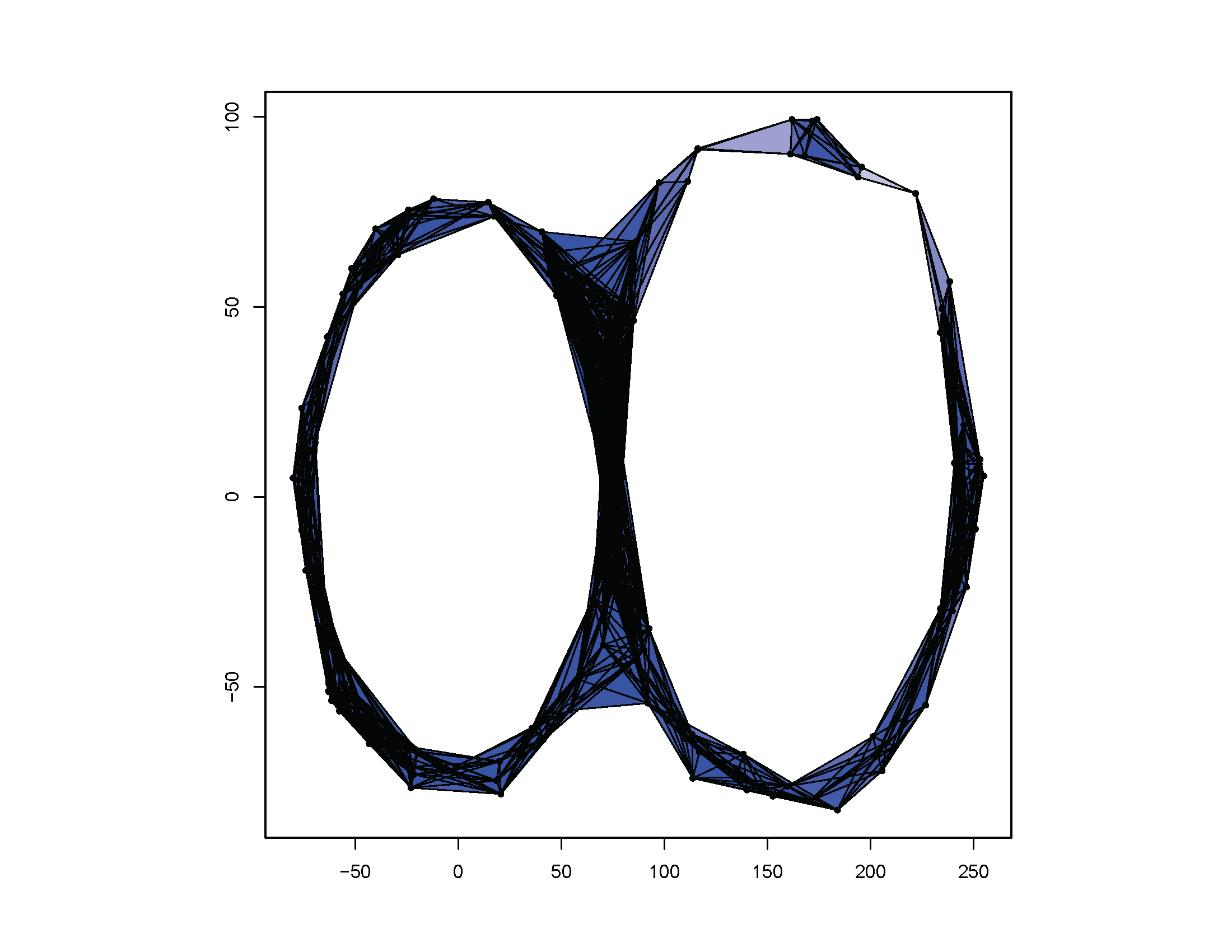}
 \includegraphics[scale=0.14]{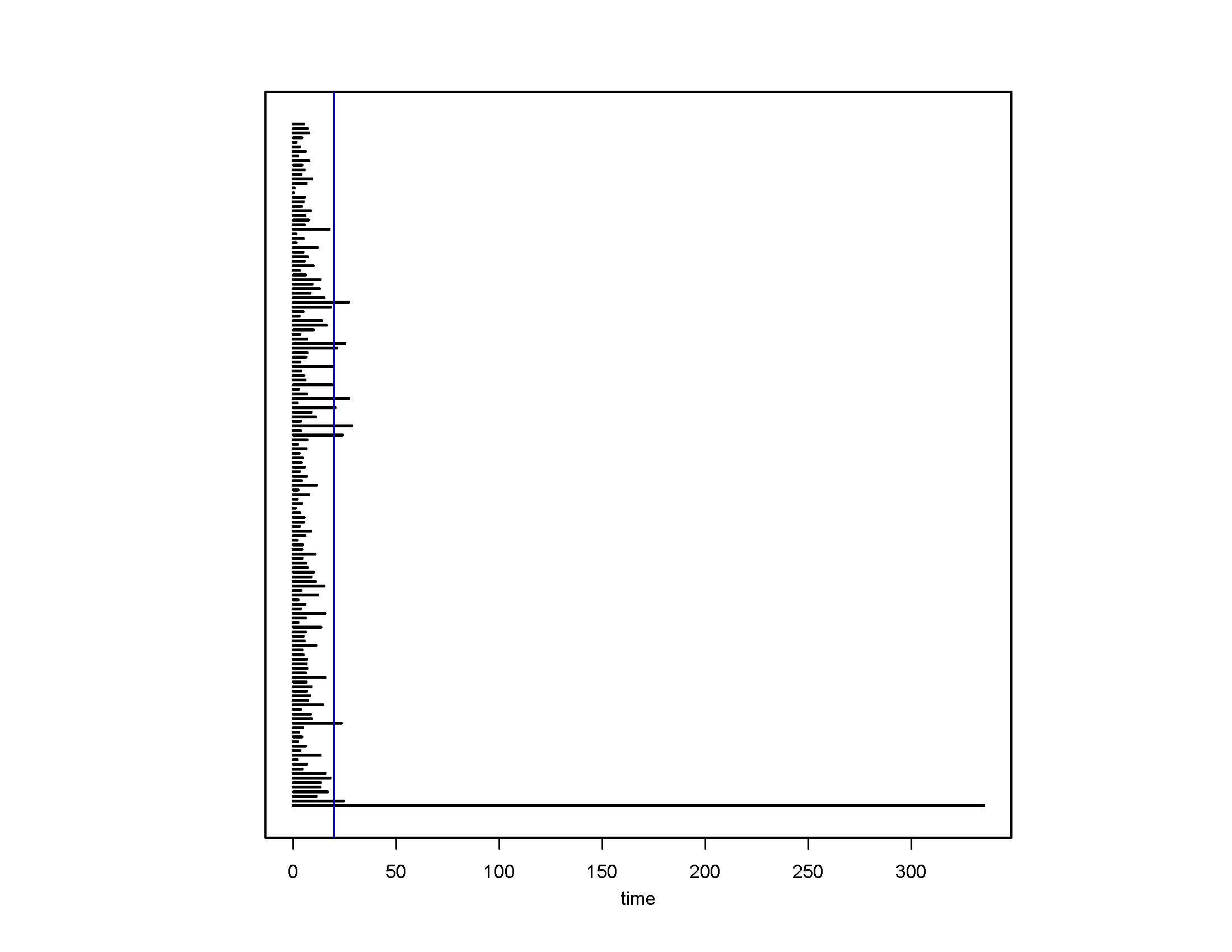}
 \includegraphics[scale=0.14]{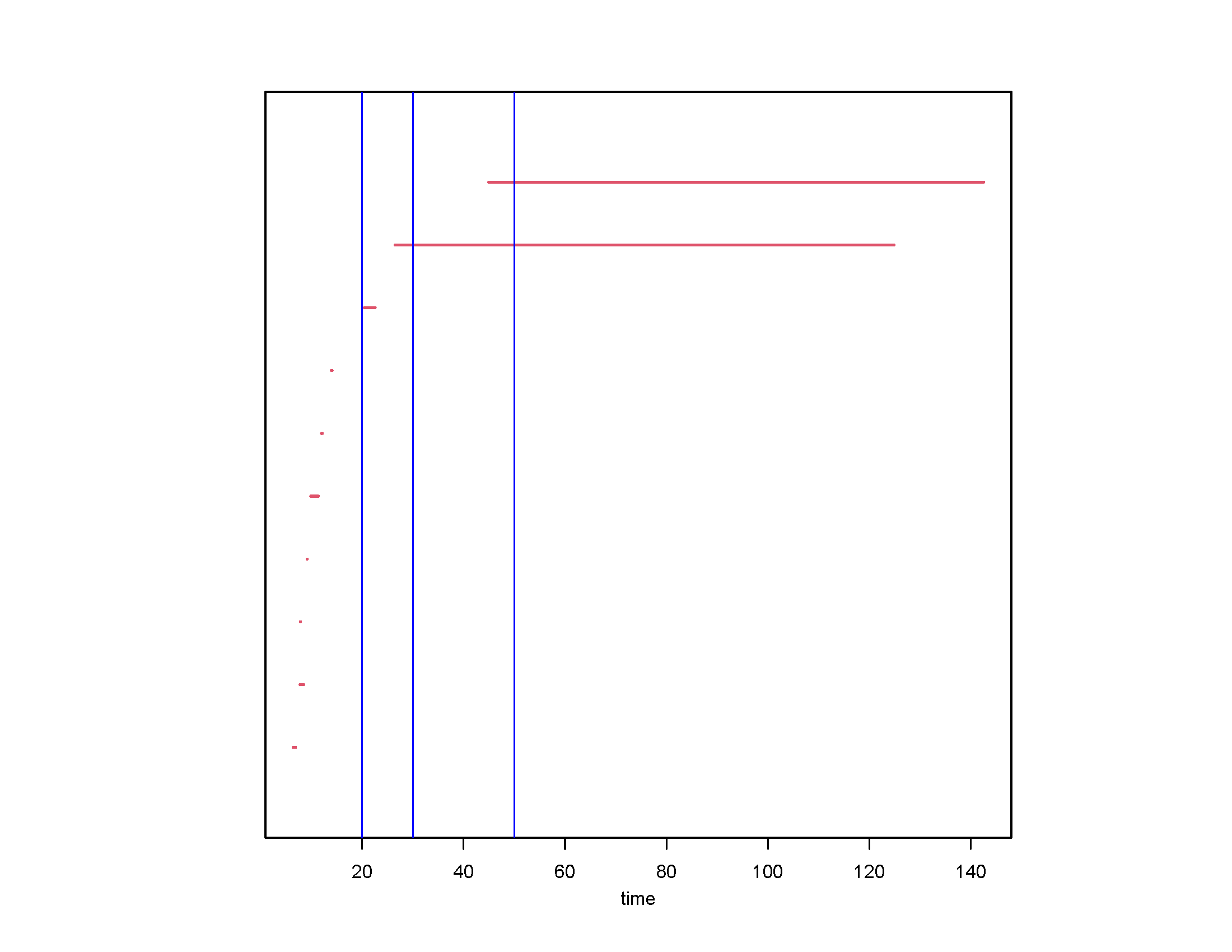}
    \caption{ Depict of evolution of Vietoris-Rips filtration. (Top left) Point cloud 150 randomly generated from  an infinity symbol~($\infty$). (Top middle) Rips complex at $\eps=20.$ (Top right) Rips complex at $\eps=30.$ (Bottom left) Rips complex at $\eps=50.$ (Bottom middle) Dimension-zero barcode with vertical line at $\eps=20.$
    (Bottom right) Dimension-one barcode with vertical lines indicating $\eps=20, 30$ and  $50.$}
    \label{fig:ripsfilt}
\end{figure}

In Figure \ref{fig:ripsfilt}, the persistent homology of a Vietoris-Rips complex is illustrated with 150 random samples from a infinity symbol~($\infty$) in $\real^2$.  The number of connected components and loops in the complex are referred to as the first and second betti numbers, denoted $\beta_0$ and $\beta_1$ respectively. In general, $\beta_p$ counts the number of $p$-dimensional holes in the complex. If a homological feature persists for a long time, it is considered a true feature, meanwhile features that appear and disappear quickly are considered noise. 

\begin{figure}[h!]
\begin{center}
\begin{tabular}{cccc}
\includegraphics[scale=.30]{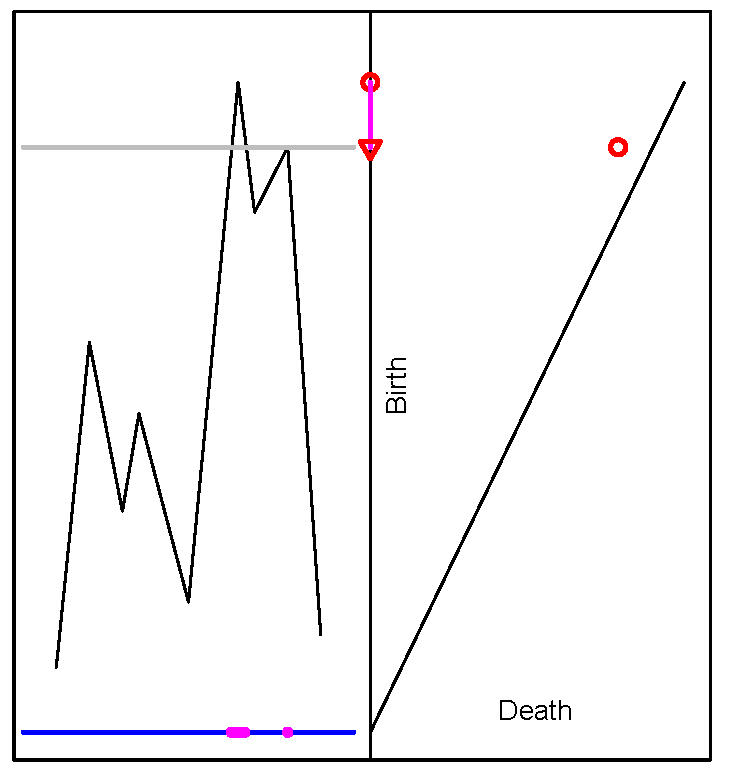}
\includegraphics[scale=0.30]{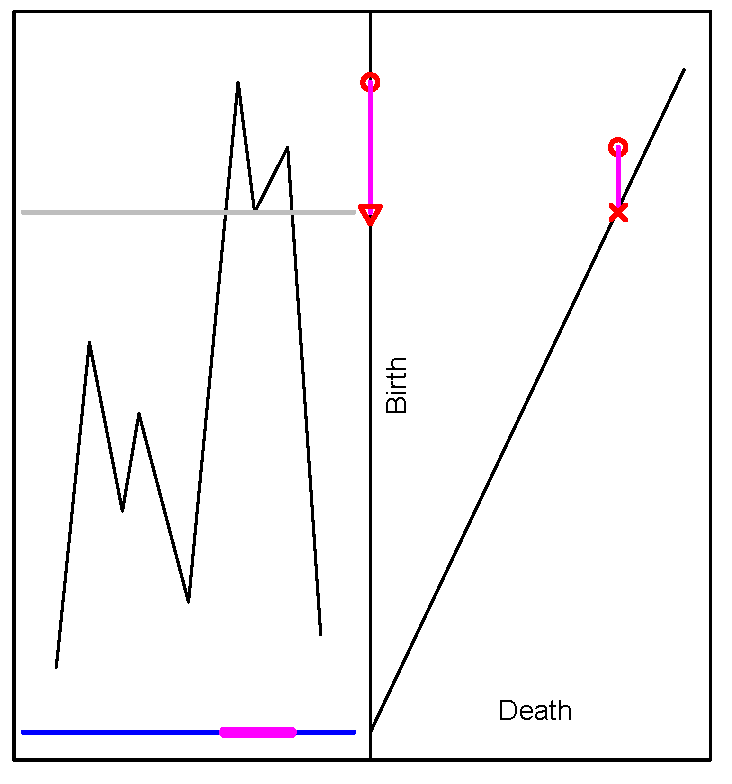} 
\includegraphics[scale=0.30]{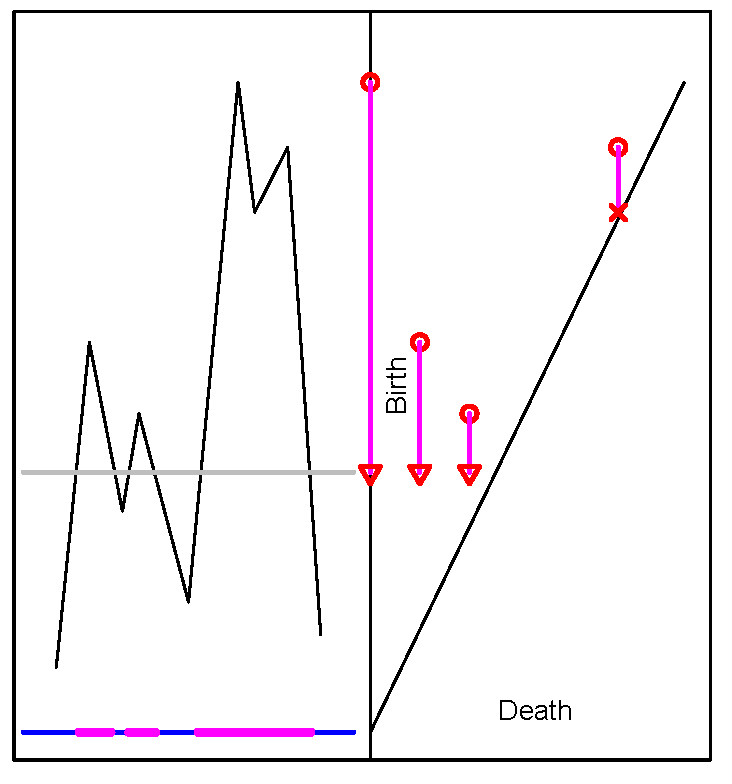} \includegraphics[scale=0.30]{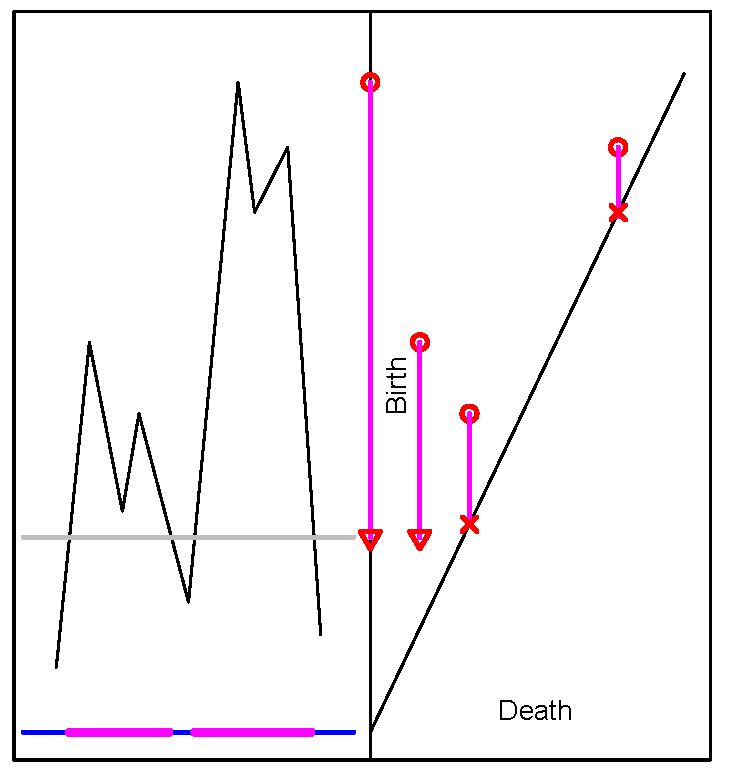}\\
\end{tabular}
\end{center}
%\vskip0.1cm
\caption{(From left to right) (1) One component is born at a local maximum ($\circ$), persists, and a new component is born at another local maximum ($\circ$). (2) The components which appeared at the local maxima have merged at a local minimum ($\times $), thus $\beta_0=1$. 
(3) Two new components that were born at local maxima and the same component from (2) persist ($\beta_0=3$).
(4) While there are two short-lived components, two components persist ($\beta_0 = 2$):
it is therefore likely that the data set is sampled from a bimodal distribution.}
\label{fig:morseF}
\end{figure}

In Figure \ref{fig:morseF}, we illustrate the super-level subset filtration of a synthetic discretized Morse function.

\begin{figure}[h!]
\centering
 \includegraphics[scale=0.14]{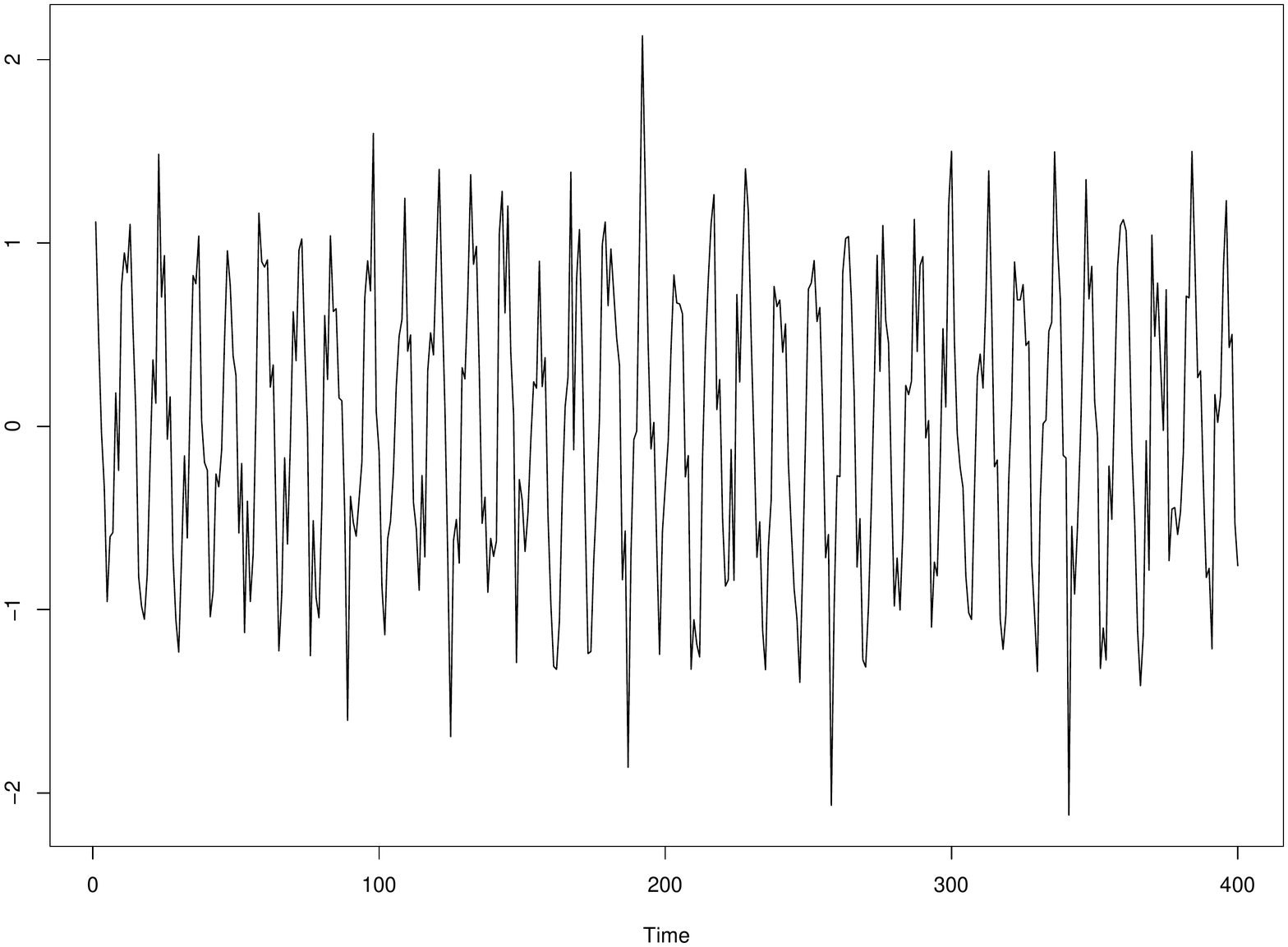}  \includegraphics[scale=0.41]{ImagesII/newEx1Taken3D_snapshot.pdf} \includegraphics[scale=0.15]{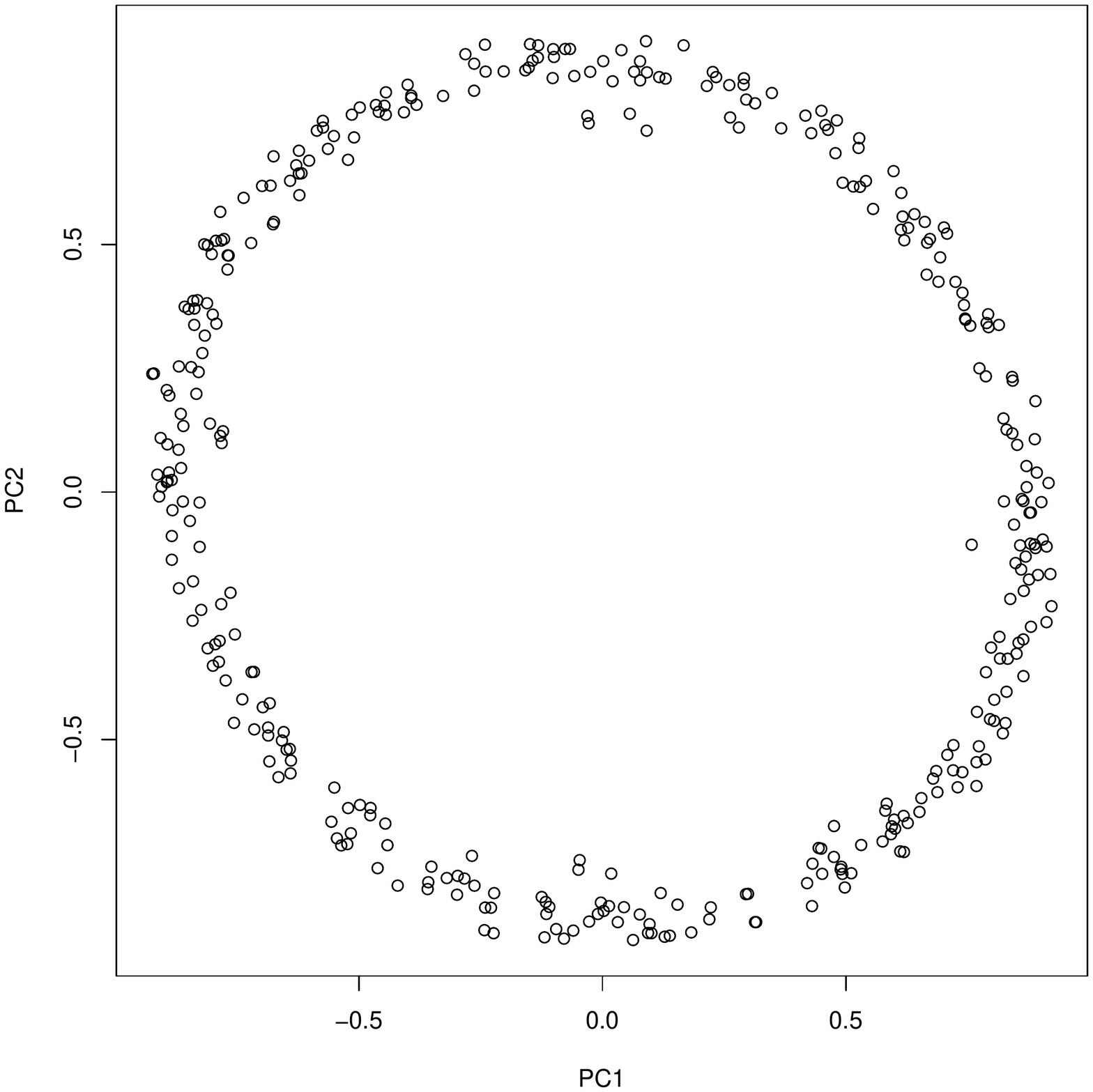}\\
 \includegraphics[scale=0.14]{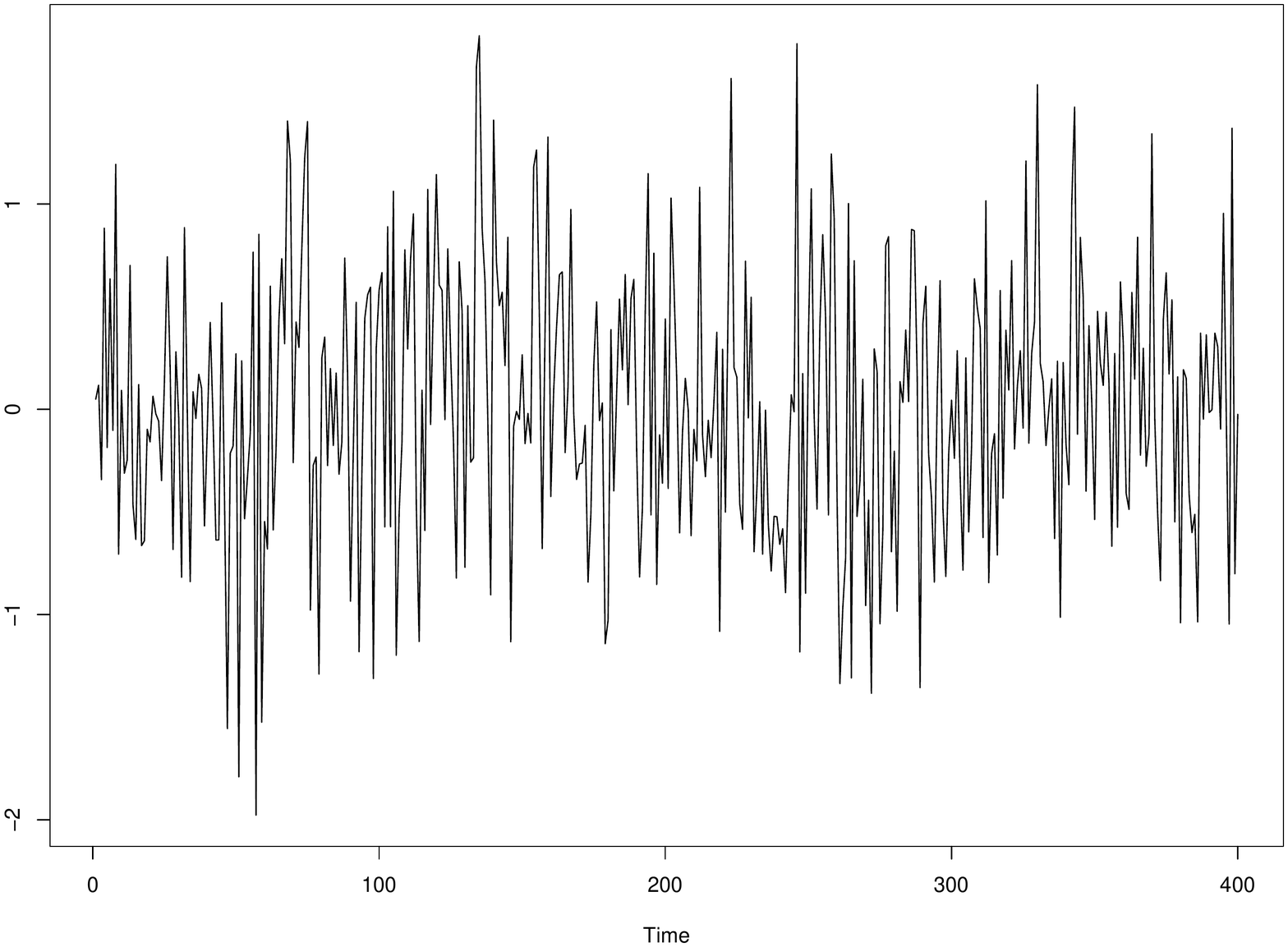}
 \includegraphics[scale=0.41]{ImagesII/newEx2Taken3D_snapshot.pdf}
 \includegraphics[scale=0.14]{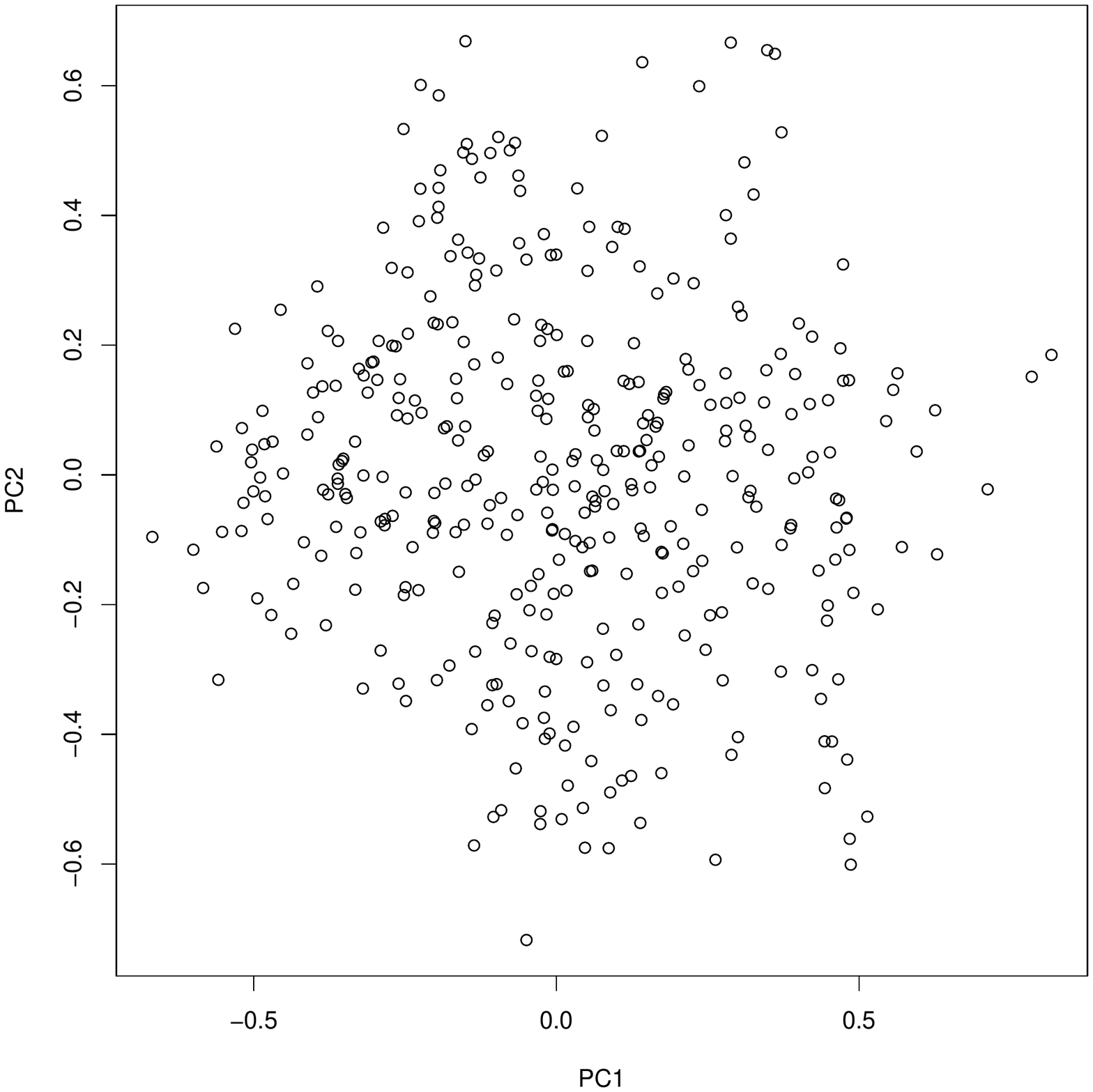}
 \caption{Transforming time series to point cloud: (top left) synthetic  time series (periodic). (top middle) point cloud embedded in 3D by sliding window. (top right) point cloud in 2D, plot PC1 vs. PC2. (bottom left) white noise (non-periodic). (top middle) point cloud embedded in 3D by sliding window. (top right) point cloud in 2D, plot PC1 vs. PC2.}
\label{fig:embedding}
\end{figure}

In Figure \ref{fig:embedding}, we illustrate the sliding window embedding with two time series.
The time series on the top row is generated from $X(t)=cos(2\pi t/12)+w(t),$ where $w(t)$ white noise $\sim N(0, 0.36)$.
The time series at the second row is white noise $w(t) \sim N(0, 0.64).$we can see that PCA of point cloud shows the periodicity, while white noise does not.

%\section*{References}
%Here are two sample references: %\cite{Feynman1963118,Dirac1953888}.
%\bibliography{mybibfile}
\end{document}